\documentclass[english,a4paper,11pt]{article}
\usepackage{fullpage}

\usepackage{amsxtra, amsfonts, amssymb, amstext}
\usepackage{amsthm}
\usepackage{booktabs}
\usepackage{longtable}
\usepackage{nicefrac}
\usepackage{balance}
\usepackage{xspace}
\usepackage[noadjust]{cite}
\usepackage{url}
\usepackage{graphics,color}
\usepackage[colorlinks]{hyperref}
\definecolor{linkblue}{rgb}{0.1,0.1,0.8}
\hypersetup{colorlinks=true,linkcolor=linkblue,filecolor=linkblue,urlcolor=linkblue,citecolor=linkblue}
\usepackage[algo2e,ruled,vlined,linesnumbered]{algorithm2e}
\newcommand{\assign}{\leftarrow}
\usepackage{graphicx}
\usepackage{wrapfig}

\usepackage{pgfplots}
\usepackage{pgfplotstable}
\usetikzlibrary{spy}

\usepackage{subfig}
\usepackage{flushend}
\usepackage{longtable}
\usepackage{rotating} 

\newcommand{\ignore}[1]{}

\allowdisplaybreaks[1]

\newcommand{\R}{\mathbb{R}}

\renewcommand{\epsilon}{\varepsilon}

\newcommand{\onemax}{\textsc{OneMax}\xspace}

\newcommand{\OneMax}{\textsc{OneMax}\xspace}
\newcommand{\OM}{\textsc{Om}\xspace}

\newcommand{\leadingones}{\textsc{LeadingOnes}\xspace}

\newcommand{\oea}{$(1 + 1)$~EA\xspace}
\newcommand{\oeaab}{$(1 + 1)~\text{EA}(A,b)$\xspace}

\newcommand{\opl}{$(1 + \lambda)$~EA\xspace}
\newcommand{\lea}{$(1 + \lambda)$~EA\xspace}

\newcommand{\AB}{$(1+\lambda)$~EA$(A,b)$\xspace}
\newcommand{\ABlb}{$(1+\lambda)$~EA$(A,b,1/n^2)$\xspace}
\newcommand{\rate}{2-rate $(1+\lambda)$~EA$_{r/2,2r}$\xspace}
\newcommand{\rateC}{2-rate $(1+\lambda)$~EA$_{c_1r,c_2r}$\xspace}
\newcommand{\ratelb}{2-rate $(1+\lambda)$~EA$_{r/2,2r}(1/n^2)$\xspace}
\newcommand{\threerate}{3-rate $(1+\lambda)$~EA$_{r/2,r,2r}$\xspace}
\newcommand{\threerateC}{3-rate $(1+\lambda)$~EA$_{c_1r,r,c_2r}$\xspace}
\newcommand{\oplshift}{$(1 + \lambda)$~EA$_{0\rightarrow 1}$\xspace}

\DeclareMathOperator{\mut}{flip}
\DeclareMathOperator{\Bin}{Bin}

\begin{document}
\title{Offspring Population Size Matters when Comparing Evolutionary Algorithms with Self-Adjusting Mutation Rates}
 
\author{Anna Rodionova$^1$, Kirill Antonov$^1$, Arina Buzdalova$^1$, Carola Doerr$^2$}
\date{$^1$ITMO University, Saint Petersburg, Russia\\
$^2$Sorbonne Universit\'e, CNRS, LIP6, Paris, France}

\maketitle

\begin{abstract}
We analyze the performance of the 2-rate $(1+\lambda)$ Evolutionary Algorithm (EA) with self-adjusting mutation rate control, its 3-rate counterpart, and a $(1+\lambda)$~EA variant using multiplicative update rules on the OneMax problem. We compare their efficiency for offspring population sizes ranging up to $\lambda=3,200$ and problem sizes up to $n=100,000$.

Our empirical results show that the ranking of the algorithms is very consistent across all tested dimensions, but strongly depends on the population size. While for small values of $\lambda$ the 2-rate EA performs best, the multiplicative updates become superior for starting for some threshold value of $\lambda$ between 50 and 100. Interestingly, for population sizes around 50, the $(1+\lambda)$~EA with static mutation rates performs on par with the best of the self-adjusting algorithms. 

We also consider how the lower bound $p_{\min}$ for the mutation rate influences the efficiency of the algorithms. We observe that for the 2-rate EA and the EA with multiplicative update rules the more generous bound $p_{\min}=1/n^2$ gives better results than $p_{\min}=1/n$ when $\lambda$ is small. For both algorithms the situation reverses for large~$\lambda$.
\end{abstract}

\sloppy{
\section{Introduction}%
\label{sec:introduction}

A key driver for the success of evolutionary algorithms (EAs) is their \emph{global search behavior}, i.e., their capability of searching the whole decision space without getting stuck in local optima. This feature distinguishes EAs from other well-known heuristics such as Simulated Annealing and other local hill climbers, which search the decision space only within a small neighborhood. The global search behavior of EAs, however, comes at the cost of a less focused search when the optimization converges, which can result in performances losses in the final parts of the optimization process. The question how to most effectively combine the best of both worlds is the driving force behind research on \emph{parameter control}~\cite{KarafotiasHE15,AletiM16,EibenHM99}, \emph{adaptive operator selection}~\cite{Maturana2010,FialhoCSS10}, and \emph{hyper-heuristics}~\cite{BurkeGHKOOQ13}, which are the most prominent umbrella terms for adjusting the structure of the search behavior to the current needs of an iterative optimization process. 

Interestingly, research on parameter control has shown that in many applications the search behavior can be adjusted very efficiently by quite simple update rules, see the above-cited surveys for details. The considerable performance gains observed in practice have inspired a whole series of theoretical works on non-static parameter choices. In the last years, an increasing number of results appeared which rigorously quantify the advantages of parameter control, see~\cite{DoerrD18chapter} for a summary and classification of the mechanisms. 

We will focus in this work on two different self-adjusting approaches previously shown to yield excellent performances on the \onemax benchmark problem: a 2-rate success control and a generalized one-fifth success rule. We analyze their efficiency when implemented in a \lea framework. Our key research question concerns the scalability of performance with respect to the offspring population size~$\lambda$ and with respect to the problem dimension~$n$. A summary of findings will be presented in Section~\ref{sec:our}. 

\subsection{Self-Adjusting Algorithms}\label{sec:intro:algos}

\textbf{The One-Fifth Success Rule Applied to the \texorpdfstring{\opl}{(1+L) EA}.} The \emph{one-fifth success rule} originally stems from theoretical observations made by Rechenberg for the optimal step-size adaptation in the (1+1) Evolution Strategy~\cite{Rechenberg}. An interpretation of this rule which is suitable also for the adaptation of other parameters, such as the mutation rate in discrete EAs, was presented in~\cite{KernMHBOK04}. The rule itself is simple: if after one iteration of an algorithm a strictly better offspring has been found, the parameter under consideration is multiplied with some constant $F$, and it is multiplied with $F^{-1/4}$ otherwise. With this setting it holds that after 5 iterations the parameter is the same as in the first if exactly one out of the five iterations was ``successful'' (i.e., found a better solution), and it is increased or decreased otherwise, depending on the sign of $1-F$ and the number of successful iterations. The one-fifth success rule was shown to yield very efficient optimization times for the $(1+(\lambda,\lambda))$~Genetic Algorithm (GA), first by empirical means~\cite{DoerrDE15} and later by rigorous mathematical running time analysis~\cite{DoerrD18ga}. 

Since the one-fifth success rule had been derived only for the (1+1)~ES and only for particular benchmark problems (the sphere and a corridor function~\cite{Rechenberg}), it seems natural to generalize the update mechanisms to other success rules. In fact, multiplying the parameters by $2$ and $1/2$ in the case of a successful and an unsuccessful iteration, respectively, has been experimented with in various context. It has also been rigorously proven to be efficient for the optimization of different benchmark functions with a \opl variant that uses a self-adjusting offspring population size~$\lambda$~\cite{LassigS11}. In~\cite{DoerrW18} an empirical study analyzed the impact of the update factors on the performance of a (1+1)~EA variant. This algorithm samples in each iteration one offspring $y$ from the current best solution $x$ and updates the mutation rate $p$ to $bp$ (with $b<1$ being some constant) if $y$ is at least as good as its parent (i.e., in our setting with \emph{maximization} as objective, if and only if $f(y) \ge f(x)$). In this situation $x$ is replaced by $y$. If $y$ is strictly worse than $x$ (i.e., if $f(y)<f(x)$) the mutation rate is increased to $Ap$, where $A>1$ is again some constant. The results presented in~\cite{DoerrW18} show that this self-adjusting \oeaab is very efficient on two classic benchmark problems \onemax and \leadingones, and this holds for broad ranges of update strengths~$A$ and~$b$. In the very recent work~\cite{DoerrDL19} it was rigorously proven that this algorithm, for suitably chosen hyper-parameters $A$ and $b$, achieves optimal optimization time on \leadingones, up to lower order terms. 

In this work, we will extend the \oeaab to the \opl, which samples $\lambda$ offspring per each iteration. Since the probability of creating individuals which are equally good, but not strictly better than the parent, increases considerably for large $\lambda$, we have to refine the interpretation of ``success'' for this context. Taking into account that the fraction of offspring $y$ satisfying $f(y) \ge f(x)$ showed promising performances in initial experiments, in our \AB we therefore discriminate with respect to whether or not at least $5\%$ of the offspring are at least as good as their parent. 

\textbf{2-Rate and 3-Rate Update Rules.} An alternative way to control the mutation rate was proposed and analyzed in~\cite{DoerrGWY19}. Their 2-rate \lea, which we refer to as \rate in the following, uses two different mutation rates in each iteration. Half the offspring are created with mutation rate $p/2$ and the other $\lambda/2$ offspring are sampled with mutation rate $2p$. The mutation rate is parametrized as $p=r/n$ in the \rate. The value of $r$ is updated after each iteration, by a random decision which favors the rate by which the best offspring has been created (details will be presented in Section~\ref{sec:algorithms}). It was proven in~\cite{DoerrGWY19} that the \rate with $\lambda \ge 45$ and $\lambda=n^{O(1)}$ is not only more efficient for the optimization of \onemax than any \lea with static mutation rates, but, from a comparison with a lower bound proven in~\cite{BadkobehLS14}, their result also implies that the \rate even achieves asymptotically optimal expected optimization time, which is $\Theta\left(\frac{n}{\log \lambda} + \frac{n \log n}{\lambda}\right)$ when measured in terms of generations. 
  
We also study a 3-rate variant of \rate, which creates one third of the offspring with mutation rate $c_1r/n$, $r/n$, and $c_2r/n$, respectively, where $0<c_1<1$ and $1<c_2$ are hyper-parameters set by the user. In the original work~\cite{DoerrGWY19} a similar 3-rate variant has been studied by empirical means. We extend their preliminary study by considering 100 different pairs of $(c_1,c_2)$, but also by considering its performance for a broader range of population sizes $\lambda$ and much larger dimensions $n$.	
		
\subsection{Summary of Results}
\label{sec:our}

We investigate the efficiency of the three above-mentioned algorithms (i.e., the \AB, the \rate, and the \threerate) on \onemax, for various population sizes $\lambda$ up to 3,200 and problem sizes up to $n=100{,}000$. Their performances are compared among each other and to the traditional $(1+\lambda)$~EA using static mutation rates. 

Interestingly, the ranking of the algorithms is very consistent across all tested dimensions, and only depends on the offspring population size~$\lambda$. More precisely, we show that for small population sizes $\lambda \le 50$, the \rate performs best, while for $\lambda \ge 100$  the \AB is the most efficient among the four algorithms.  
It is also worth mentioning that for $\lambda$ of about 50, the conventional \lea with static mutation rates performs on par with the best performing self-adjusting algorithm, which is \rate. 

All our algorithms are implemented with the \emph{shift mutation operator} discussed in~\cite{CarvalhoD17}. This operator, unlike the unconditional standard bit mutation operator traditionally studied in the theory of evolutionary computation, ensures that offspring differ from their parent by at least one bit. This is achieved as follows. The operator first draws the \emph{mutation strength}~$\ell$ (i.e., the number of bits that are flipped to create the offspring) from the binomial distribution $\Bin(n,p)$, where the parameter $p$ denotes the mutation rate. While the traditional standard bit mutation operator allows $\ell=0$, it is easily seen that, for $(1+\lambda)$~EAs, which only evolve a single solution, an offspring that is identical to its parent cannot advance the search. The shift mutation operator therefore interprets $\ell=0$ as a vote for a small search radius, and flips exactly one bit, i.e., effectively using $\ell=1$. Put differently, this operator always flips at least one bit, and its probability of flipping exactly one bit equals $\Bin(n,p)(0)+\Bin(n,p)(1)$. 

When the mutation rate $p$ converges to zero, the shift mutation operator converges against the operator using mutation strength one deterministically, thus effectively reducing the global search property of standard bit mutation to a purely local search. As explained above, we would like to avoid performing a purely random search only, and therefore cap the mutation rate at a lower bound $p_{\min}.$ This lower bound can have a significant impact on the performance, as our further experimental results demonstrate. More precisely, we observe that for smaller population sizes a lower bound of~$1/n^2$ seems to work better than a lower bound of~$1/n$. Interestingly, the situation reverses for larger population sizes. These observations are true for all considered self-adjusting algorithms, but with different values of $\lambda$. For \rate, the transition population size is between $\lambda = 100$ and $\lambda = 200$, while for \AB it is between $\lambda = 400$ and $\lambda = 800$.

In light of the consistent behavior across all tested dimensions, we are confident that our findings will inspire future rigorous theoretical analyses for the self-adjusting algorithms investigated herein.



\subsection{Related Work}
Our work can, to some extent, be seen as an extension of~\cite{DoerrYRWB18}, where different \opl variants have been studied on the two benchmark problems \onemax and \leadingones. Note though that we consider here much larger offspring population sizes $\lambda$ (up to $3,200$) and much larger dimensions (up to $100{,}000$), whereas the results presented in~\cite{DoerrYRWB18} are restricted to settings with $n \le 4000$, and $\lambda \in \{1, 2, 5,10,50\}$. As discussed above, the ranking of the algorithms strongly depends on the size of $\lambda$, so that our work gives a much more complete picture than the results presented in~\cite{DoerrYRWB18}. 

\subsection{Structure of the Paper}

The rest of the paper is structured as follows. In Section~\ref{sec:algorithms} we describe the considered algorithms in more detail. Section~\ref{sec:bydim} gives a general overview of how each algorithm performs across different dimensions and population sizes. In Section~\ref{sec:lb} the influence of the lower bound for the mutation probability is studied, and conclusions about the ranking of the self-adjusting algorithms are made. Section~\ref{sec:fb} provides insights into the anytime performance of the algorithms from the fixed-budget perspective. In Section~\ref{sec:3rate} we compare the \rate with its 3-rate counterpart. Conclusions and avenues for future work are presented in Section~\ref{sec:conclusions}.


\section{OneMax and Three \texorpdfstring{$(1+\lambda)$}{(1+L)} {EA Variants}}
\label{sec:algorithms}

In this section we describe the baseline \oplshift algorithm and its two self-adjusting variants. The description of the algorithms assumes the \emph{maximization} of a pseudo-Boolean function $f: \{0,1\}^n \rightarrow \R$ as the optimization task. All empirical results in the further sections are obtained for the \OneMax benchmark problem $\OM:\{0,1\}^n \to \R, x \mapsto \sum_{i=1}^n{x_i}$. \onemax is the most widely studied benchmark problem in the theory of EAs~\cite{Jansen13,AugerD11}, but also serves as a recurring benchmark problem in empirical studies, and in particular in the context of parameter control~\cite{FialhoCSS08,Back92,Back93,JansenJW05,DoerrW18}. A discussion of properties that make \onemax a suitable test problem for adaptive algorithms was offered by Thierens in~\cite{Thierens09}. Apart from the better comparison with existing results, the availability of theoretical performance limits of adaptive and static evolutionary algorithms makes \onemax a particularly appealing benchmark problem; see~\cite{Doerr18BBC} for a survey of such \emph{black-box complexity} bounds. In the context of our work, the asymptotic $\Theta\big(\frac{n}{\log \lambda} + \frac{n \log n}{\lambda}\big)$ bound for all $\lambda$-parallel EAs proven in~\cite{BadkobehLS14} and the precise $n \ln(n) -cn \pm o(n) $ bound for any unary unbiased black-box algorithm from~\cite{DoerrDY16} are the probably most relevant results. As mentioned at the end of Section~\ref{sec:our}, we are also confident that further advances in running time analysis will allow us to convert our empirical findings into rigorous mathematical statements.

Before we describe our algorithms, we recall from~\cite{JansenJW05} that for the optimization of \onemax the best static offspring population size is $\lambda=1$. All larger values result in worse expected optimization times. However, the \emph{parallel optimization time}, measured by the number of generations needed to find the optimum, of a \opl with $\lambda>1$ can (and typically is) smaller than that of the \oea.

 \begin{algorithm2e}[t]%
	\textbf{Initialization:} 
	Sample $x \in \{0,1\}^{n}$ u.a.r. and evaluate $f(x)$\;
  \textbf{Optimization:}
	\For{$t=1,2,3,\ldots$}{
		\For{$i=1,\ldots,\lambda$}{
			Sample $\ell^{(i)}$ from $\Bin_{0 \rightarrow 1}(n,p)$, sample 
			$y^{(i)} \assign \mut_{\ell^{(i)}}(x)$ and evaluate $f(y^{(i)})$\;
		}
		\label{line:selectionopl}
		Sample $x^*$ from $\arg\max\{f(y^{(1)}), \ldots, f(y^{(\lambda)})\}$ u.a.r.\;
		\lIf{$f(x^*)\ge f(x)$}{$x \assign x^*$}	
	}
\caption{The \oplshift with mutation rate $0<p<1$ for the maximization of $f:\{0,1\}^n \rightarrow \R$}
\label{alg:oplres}
\end{algorithm2e}

The basic algorithm for our study is the \oplshift. Its description is given in Algorithm~\ref{alg:oplres}. The \oplshift samples $\lambda$ solution candidates in every iteration, but only the best one of these \emph{offspring} survives -- ties broken uniformly at random (u.a.r.). Each offspring is created by the \emph{shift mutation operator}, 
which first samples a mutation strength~$\lambda$ from the $\Bin_{0 \rightarrow 1}(n,p)$ distribution and then flips the entries in $\ell$ randomly chosen, pairwise different bit positions. Here and in the following, the distribution $\Bin_{0 \rightarrow 1}(n,p)$ assigns to each integer $2\le k\le n$ the value $\Bin_{0 \rightarrow 1}(n,p)(k)=\Bin(n,p)(k)$, sets $\Bin_{0 \rightarrow 1}(n,p)(0)=0$, and $\Bin_{0 \rightarrow 1}(n,p)(1)=\Bin(n,p)(1)+\Bin(n,p)(0)$.

We implemented sampling from $\Bin_{0 \rightarrow 1}(n,p)$ as follows. First, each bit of a string of length $n$ is inverted with probability $p$. Then it is checked whether at least one bit was inverted. If not, one random bit position is chosen uniformly at random and the corresponding bit is inverted.

We note without going into much detail that apart from the ``shift'' strategy $0 \to 1$, which assigns the probability mass of sampling $\ell=0$ to $\ell=1$, a second approach to deal with 0-bit flips was suggested in~\cite{CarvalhoD17}. This alternative approach, coined ``resampling strategy'' and denoted by a subscript $>0$ in~\cite{CarvalhoD17} and its follow-up works~\cite{DoerrYRWB18,DoerrW18}, distributes the probability mass $\Bin(n,p)(0)$ proportionally to all integers $1 \le \ell \le n$. Initial experiments with this resampling strategy indicate similar finding as those presented for the shift strategy used here in this work. A detailed examination is left for future work. 

Next, we consider two self-adjusting variants of the \lea. The first one is the \rate proposed in~\cite{DoerrGWY19}. This algorithm is summarized in Algorithm~\ref{alg:DoerrGWY19}. It creates one half of the offspring with mutation rate $2r/n$ and the other half of the offspring with mutation rate $r/(2n)$, respectively. At the end of each iteration, the \rate updates the mutation rate based on which value was used when the best offspring was obtained. Note here that with probability $1/2$ a random decision is made whether to update the mutation rate to $2r/n$ or to $r/(2n)$, thus effectively assigning a $3/4$ chance to update to the value by which the best offspring has been created. The reason not to update to this rate deterministically was explained in~\cite{DoerrGWY19} by the fact that the constant probability to update the mutation rate in the seemingly unfavorable direction can be useful in the optimization of non-unimodal functions; see~\cite[Section~3.1]{DoerrGWY19} for a more detailed discussion.

 \begin{algorithm2e}[t]%
	\textbf{Initialization:} 
	Sample $x \in \{0,1\}^{n}$ u.a.r. and evaluate $f(x)$\;
	Initialize $r \assign r^{\text{init}}$; // Following~\cite{DoerrGWY19} we use $r^{\text{init}}=2$\;
  \textbf{Optimization:}
	\For{$t=1,2,3,\ldots$}{
		\For{$i=1,\ldots,\lfloor\lambda/2\rfloor$}{
			\label{line:mut}Sample $\ell^{(i)}$ from $\Bin_{0\rightarrow 1}(n,r/(2n))$, 
			create $y^{(i)} \assign \mut_{\ell^{(i)}}(x)$, and
			evaluate $f(y^{(i)})$\;
		 }
		\For{$i=\lfloor\lambda/2\rfloor+1,\ldots,\lambda$}{
			\label{line:double}Sample $\ell^{(i)}$ from $\Bin_{0 \rightarrow 1}(n,2r/n)$, 
			create $y^{(i)} \assign \mut_{\ell^{(i)}}(x)$, and
			evaluate $f(y^{(i)})$\;
		 }
		$x^* \assign \arg\max\{f(y^{(1)}), \ldots, f(y^{(\lambda)})\}$ (ties broken u.a.r.)\;
		\lIf{$f(x^*)\ge f(x)$}{$x \assign x^*$}
		Perform one of the following two actions with prob. $1/2:$
        \begin{itemize}
           \item replace $r$ with the mutation rate that $x^*$ has been created with;
           \item replace $r$ with either $2r$ or $r/2$ equiprobably.
        \end{itemize}  
        $r \assign \min\{\max\{2, r\}, n/4\}$\;
	}
\caption{The \rate with adaptive mutation rates proposed in~\cite{DoerrGWY19}}
\label{alg:DoerrGWY19}
\end{algorithm2e}  

We complement our study by a performance analysis for an extension of the $(1+1)$~EA$(A,b)$ studied in~\cite{DoerrW18}. In this extension we adapt the success-based multiplicative update rule used in the $(1+1)$~EA$(A,b)$ to the case that more than one offspring is generated per each iteration. We call our extension the \AB. Algorithm~\ref{alg:1LAb} provides its pseudo-code. In the original algorithm from~\cite{DoerrW18}, the mutation rate is updated based on whether or not the offspring replaces its parent. In our case, we have $\lambda$ offspring, and we refine the update rule as follows. First, we compute the number of ``good'' offspring, which are at least as good as the parent (see line~\ref{line:N} of Algorithm~\ref{alg:1LAb}). If the share of good offspring is at least 5\%, the mutation rate is multiplied by the factor $A>1$. It is decreased to $bp$ otherwise, where the factor $b$ is some constant satisfying $0<b<1$. In this work we set $A=2$ and $b=1/2$. 

One may wonder why we use the 5\% threshold. This is based on some preliminary experiments with $\lambda=1{,}600$, where we observed good performances for this value. It is likely that this threshold depends on the population size $\lambda$. A detailed study is left for future work.

 \begin{algorithm2e}[h!]%
	\textbf{Initialization:} 
	Sample $x \in \{0,1\}^{n}$ u.a.r. and evaluate $f(x)$\;
	Initialize $p \assign 1/n$\;
  \textbf{Optimization:}
	\For{$t=1,2,3,\ldots$}{
		\For{$i=1,\ldots,\lambda$}{
			\label{line:half}Sample $\ell^{(i)}$ from $\Bin_{0 \rightarrow 1}(n,p)$, 
			create $y^{(i)} \assign \mut_{\ell^{(i)}}(x)$, and
			evaluate $f(y^{(i)})$\;
		 }
		$N \assign |\{ i \in [\lambda] \mid f(x^{(i)}) \ge f(x) \}|$\; \label{line:N}
		\lIf{$N \ge \lceil 0.05\lambda \rceil$}{$p \assign \min\{1/2,Ap\}$ \textbf{\text{ else}} $p \assign \max\{1/n,bp\}$} \label{line:0.05}
		$x^* \assign \arg\max\{f(x^{(1)}), \ldots, f(x^{(\lambda)})\}$ (ties broken u.a.r.)\;
		\lIf{$f(x^*)\ge f(x)$}{$x \assign x^*$}
	}
\caption{The \AB with adaptive mutation rates and update strengths $A>1$, $0<b<1$}
\label{alg:1LAb}
\end{algorithm2e}

\section{Algorithm's Performance by Dimension} 
\label{sec:bydim}

\begin{figure}[h!]
\centering
\includegraphics[width=\linewidth]{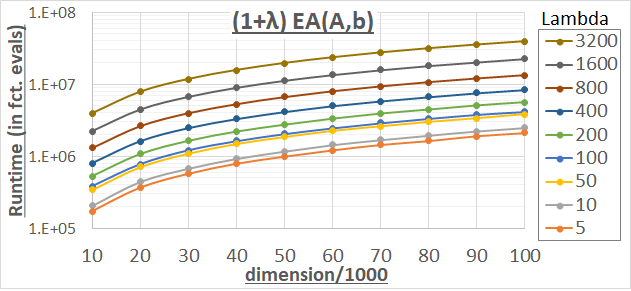}
\caption{Average optimization times for 100 independent runs of the \AB for different values $5 \le \lambda \le 3,200$ and different problem dimensions $10^4 \le n \le 10^5$. }
\label{fig:RT-AB-byLambda}
\end{figure}

\begin{figure}[h!]
\centering
\includegraphics[width=\linewidth]{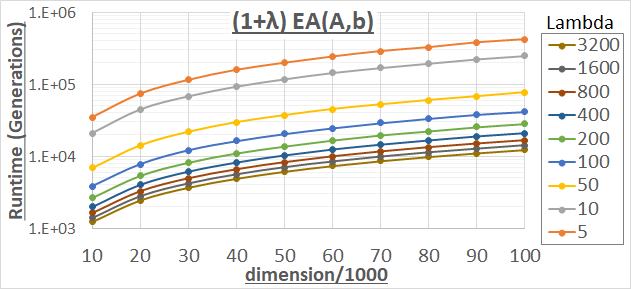}
\caption{Average parallel optimization times for 100 independent runs of the \AB for different values $5 \le \lambda \le 3,200$ and different problem dimensions $10^4 \le n \le 10^5$. }
\label{fig:RT-AB-byLambda-gen}
\end{figure}

\begin{figure*}[h!]
  \centering
  \subfloat[]{\includegraphics[width=0.5\linewidth]{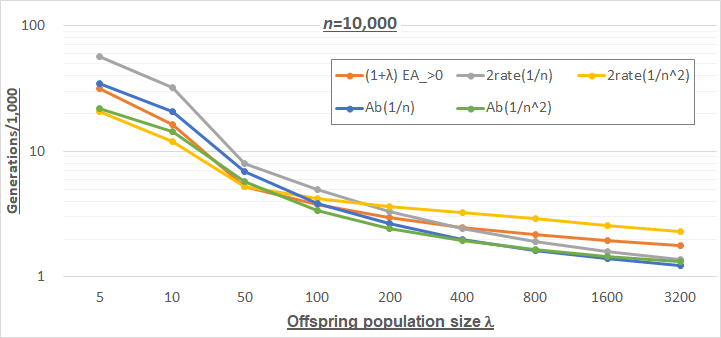}\label{fig:Gen10klog}}
  \subfloat[]{\includegraphics[width=0.5\linewidth]{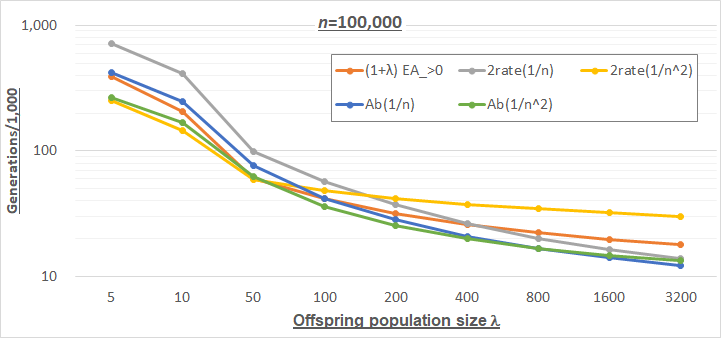}\label{fig:Gen100klog}}
   \caption{Average parallel optimization times for the different \opl variants to find an optimal solution (a) on the $10{,}000$-dimensional and (b) on the $100{,}000$-dimensional \onemax problem. The averages are for 100 independent runs each.}
  \label{fig:Genlog}
\end{figure*}

\begin{figure*}[h!]
  \centering
  \subfloat[]{\includegraphics[width=0.45\linewidth]{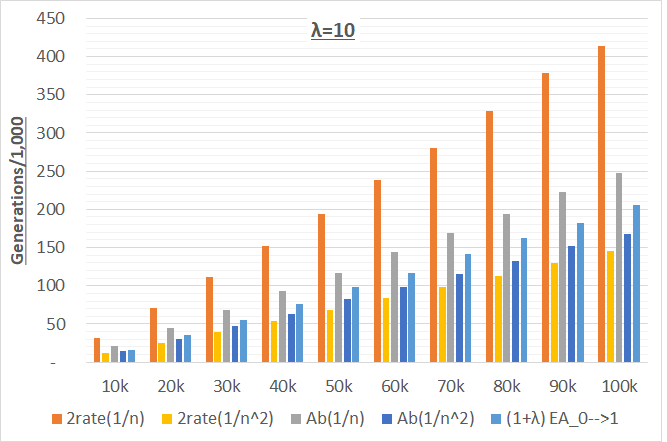}\label{fig:lambda10}}
  \subfloat[]{\includegraphics[width=0.45\linewidth]{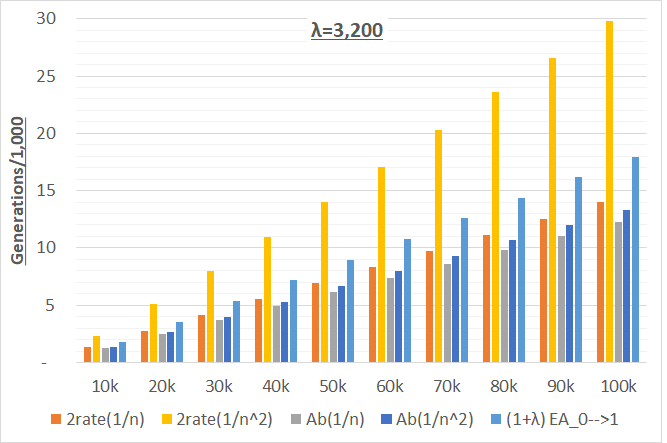}\label{fig:lambda3200}}
   \caption{Average parallel optimization times for (a) the different (1+10)~EA variants and (b) the different (1+3,200)~EA variants to find an optimal solution on the $n$-dimensional \onemax problem, with $10^4 \le n \le 10^5$. The averages are for 100 independent runs each.}
  \label{fig:lambda}
\end{figure*}

In a first step, we analyze how each algorithm performs across different problem dimensions. To this end, we compute the average optimization time (i.e., the number of function evaluations needed to find an optimal solution) of each of the algorithms described in Section~\ref{sec:algorithms} for 100 independent runs. 
In Figure~\ref{fig:RT-AB-byLambda} we show these values for the algorithm \AB, for different values of $\lambda$. We also show the average parallel optimization time, measured by the number of generations needed to find an optimal solution, for the \AB in Figure~\ref{fig:RT-AB-byLambda-gen}. The parallel optimization time equals the average optimization time divided by the offspring population size $\lambda$. This performance measure is useful when the fitness evaluations within one generation are made in parallel. 

We note without details that the plots for the \rate and \oplshift look very similar, but with different values. The relative standard deviation is small for all algorithms: it starts at about 10-15\% for $\lambda=5$ and decreases to just 0.5-1.5\% for $\lambda=3200$, so we can draw significant conclusions from the average optimization time and the average parallel optimization time plots. Exact figures and standard deviations are available in the appendix.

As expected, we see that for each algorithm and each dimension the average optimization times strictly increase with increasing $\lambda$. Therefore, small values of $\lambda$ are optimal in terms of optimization time. However, in terms of parallel optimization time, the situation is reversed: the algorithms with larger population sizes $\lambda$ require fewer iterations to find the optimum. Therefore, both small and large values of $\lambda$ are worth being used under different circumstances. 



\section{The Impact of the Lower Bound}
\label{sec:lb}

One of the key advantages of sampling the mutation rates from the distribution $\Bin_{0 \rightarrow 1}(n,p)$ is that it allows to transition from a classical \opl to a $(1+\lambda)$ variant of RLS that creates each offspring by flipping exactly one uniformly chosen bit. This transition is achieved by reducing $p$ beyond $1/n$, see~\cite{CarvalhoD18} for a discussion. In the algorithms studied above, we had enforced a lower bound of $1/n$ for $p$. We now relax this lower bound and only require that $p \ge 1/n^2$. We study the impact of this relaxed lower bound on the performance of the \AB and the \rate algorithm. The algorithms with these relaxed lower bounds are denoted by \ABlb and \ratelb, respectively.

%
%

In Figures~\ref{fig:Gen10klog} and~\ref{fig:Gen100klog} we show the average parallel optimization times of the different algorithms for $n=10{,}000$ and $n=100{,}000$, respectively. Recall that the average parallel optimization time equals the average number of generations. Note also that we use a logarithmic scale, to ease the comparison. 
Note that the standard deviation is still relatively small for the algorithms using $p_{\min}=1/n^2$ as well. The only difference is that for \ratelb it does not decrease with increasing $\lambda$ and is always around 5-10\%.

The picture for $n=10{,}000$ is quite similar to that for $n=100{,}000$, and, in fact, for all tested dimensions. To demonstrate this, we plot in Figures~\ref{fig:lambda10} and~\ref{fig:lambda3200} the average parallel optimization times for fixed $\lambda=10$ and $\lambda=3,200$, respectively, which confirm a stable ranking of the different algorithms. We use the parallel optimization time again as performance measure, to ease the comparison with Figures~\ref{fig:Gen10klog} and~\ref{fig:Gen100klog}.

From Figures~\ref{fig:Gen10klog} and~\ref{fig:Gen100klog} we can derive threshold values for $\lambda$ at which the ranking of the algorithms changes. 
\begin{itemize}
	\item For $5 \leq \lambda \leq \lambda_1$ with $\lambda_1$ being a threshold value between 50 and 100, the \ratelb performs best.
	\item For $\lambda_1 \leq \lambda \leq \lambda_2$ with $400 \le \lambda_2 < 800$ the \ABlb shows best performance.
	\item For $\lambda_2 \leq \lambda \leq 3,200$ the \AB minimizes the expected optimization time.
\end{itemize}

We see that the ranking of algorithms strongly depends on the population size $\lambda$. For small values of $\lambda,$ the \ratelb performs the best, while for medium values the \ABlb wins, and then for large values of $\lambda$ the \AB is superior. Interestingly, for $\lambda = 50$, the \oplshift, which uses a static mutation rate, performs on par with the best self-adjusting algorithm. 

Generally, for both self-adjusting algorithms, it is more preferable to use $1/n^2$ as lower bound for the mutation probability when the population size is small, while for large population sizes a lower bound of $1/n$ seems to work better. For the \rate, the transition population size is some value $\lambda_3$ satisfying $100 \le \lambda_3 < 200$, while for the \AB it is some $\lambda_2$ with $400 \le \lambda_2 < 800$. At the same time, while the performance of the \rate and the \ratelb drastically depends on the population size and the mutation probability lower bound, the \ABlb is more stable and performs rather good for all considered population sizes. Note that the \ABlb is the only self-adjusting algorithm which is never significantly worse than the baseline \lea.

\section{Fixed-Budget Performance}
\label{sec:fb}

\begin{figure}[t]
\centering
\includegraphics[width=\linewidth]{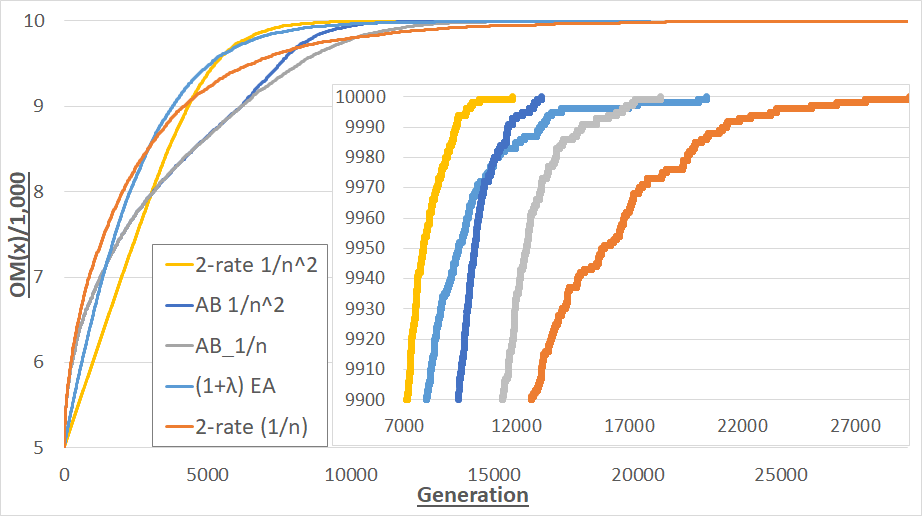}
\caption{Evolution of the average function value over 100 independent runs of different $(1+10)$~EA variants on the $10{,}000$-dimensional \onemax problem, in dependence of the generation. The small plot in the lower right corner zooms into the region in which all algorithms have found function values $\ge 9,900$. 
}
\label{fig:FB-small}
\end{figure}

\begin{figure*}[h!]
  \centering
  \subfloat[]{\includegraphics[width=0.5\linewidth]{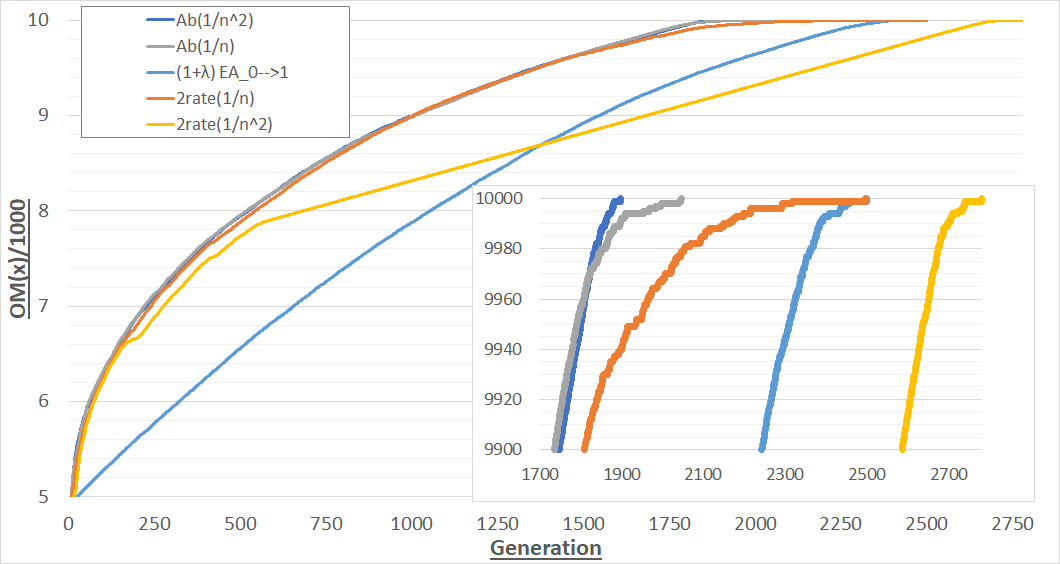}\label{fig:FB-medium}}
  \subfloat[]{\includegraphics[width=0.5\linewidth]{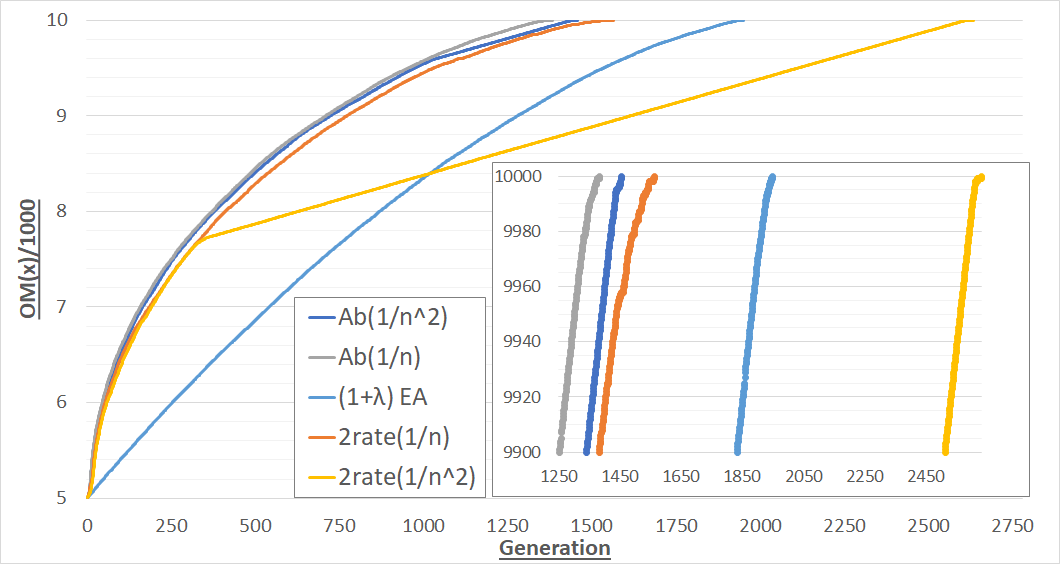}\label{fig:FB-big}}
   \caption{Evolution of the average function value over 100 independent runs of different $(1+400)$~EA (a) and $(1+1600)$~EA (b) variants on the $10{,}000$-dimensional \onemax problem, in dependence of the generation.}
  \label{fig:FB}
\end{figure*}

In the sections above, we have only regarded the average optimization times. This measure, albeit very commonly used in the runtime analysis community, does not give any information about how the algorithms perform in an \emph{anytime} sense. This criticism was addressed in~\cite{JansenZ14,CarvalhoD17}, where the benefits of \emph{fixed-budget} and \emph{fixed-target} analyses are advertised, respectively. We note that, of course, both performance measures have played an important role in empirical evaluations ever since -- the contribution of the mentioned papers is rather to be seen in discussing why these measures provide important insights for theoreticians. Figures~\ref{fig:FB-small}, \ref{fig:FB-medium}, \ref{fig:FB-big} show anytime performance plots for selected $(1+10)$~EAs, $(1+400)$~EAs, and $(1+1600)$~EAs, respectively, on the $n=10{,}000$-dimensional \onemax instance. 

We see that for $\lambda=10$ (Figure~\ref{fig:FB-small}) the algorithms demonstrate quite different performance over time, while for larger values of $\lambda$ (Figures~\ref{fig:FB-medium}, \ref{fig:FB-big}) the performance of the most efficient algorithms seems to be rather stable. 

More precisely, the situation is as follows. For generational budgets (note that the budget in terms of fitness evaluations is simply $\lambda$ times larger than the generational budget, since we are dealing with constant offspring population sizes in this paper) up to around 2{,}590 the \rate is the most efficient algorithm; then from 2{,}950 to 5{,}615 generations \oplshift is the best one, then the \ratelb, which remains to be the best performing algorithm for all budgets up to 11{,}800, at which point it hits the optimum. The runner-up is the \ABlb, third the \AB, then the \oplshift and the worst among the five algorithms is the \rate (with lower bound $p_{\min}=1/n$). Therefore, for small population sizes, the ranking of the algorithms depends on the budget. 

For $\lambda = 400$ and $\lambda = 1{,}600$, we see that for all budgets the two algorithms \ABlb and \AB show similar performance, and are the best among the five algorithms. For $\lambda=400$ the \rate shows fine performance for budgets up to around 1{,}700, at which point its performance deteriorates (due to too many offspring sampled with four times the in this case optimal mutation rate). For $\lambda=1{,}600$ this effect disappears, and the \rate shows fine performance. The performance of the \ratelb is seen to be the worst of all algorithms for budgets greater than $1{,}400$ for $\lambda=400$ and budgets larger than $1{,}000$ for $\lambda=1{,}600$.



\section{2-Rate vs. 3-Rate Adaptation}
\label{sec:3rate}

We next analyze an idea previously communicated in~\cite{DoerrGWY19}, a 3-rate success-based variant of \rate, which we call the \threerate. This algorithm creates one third of the offspring with mutation rate $c_1r/n$, $r/n$, and $c_2r/n$, respectively, where $0<c_1<1$ and $1<c_2$ are hyper-parameters set by the user. The detailed description of the \threerate is given in Algorithm~\ref{alg:threerate}. In the original work~\cite{DoerrGWY19} a similar 3-rate variant has been studied. However, there is no pseudocode of this variant, so it is hard to compare the update rule used in the original work and the one we use. 

Following~\cite{DoerrGWY19}, \emph{we use the standard bit mutation operator for all algorithms in this section}. It inverts each bit of a string equiprobably with the corresponding mutation rate. So, in contrast to the shift mutation operator used above, it may happen that no bits are flipped in the mutation phase. However, for moderately large $\lambda$, the shift mutation operator performs only slightly better than the standard one (this empirical observation is also confirmed by the findings made in~\cite{DoerrYRWB18} for the \emph{resampling} strategy, and can be confirmed theoretically; see the full version of~\cite{CarvalhoD18} for a similar theoretical justification in the context of the \lea with static mutation rates), so we believe that the key observations made in this section do not depend on the choice of the mutation operator. 

 \begin{algorithm2e}%
	\textbf{Initialization:} 
	Sample $x \in \{0,1\}^{n}$ u.a.r. and evaluate $f(x)$\;
	Initialize $r \assign r^{\text{init}}$; // We use $r^{\text{init}}=2$\;
	Initialize $c_1 \ge 1, 0 < c_2 < 1$; // We use $c_1 = 0.7, c_2 = 1.4$\;
  \textbf{Optimization:}
	\For{$t=1,2,3,\ldots$}{
		\For{$i=1,\ldots,\lfloor\lambda/3\rfloor$}{
			\label{line:first}Sample $\ell^{(i)}$ from $\Bin(n,c_1r/n)$, 
			create $y^{(i)} \assign \mut_{\ell^{(i)}}(x)$, and
			evaluate $f(y^{(i)})$\;
		 }
		\For{$i=\lfloor\lambda/3\rfloor+1,\ldots,\lfloor 2\lambda/3\rfloor$}{
			\label{line:second}Sample $\ell^{(i)}$ from $\Bin(n,r/n)$, 
			create $y^{(i)} \assign \mut_{\ell^{(i)}}(x)$, and
			evaluate $f(y^{(i)})$\;
		 }
		\For{$i=\lfloor 2\lambda/3\rfloor+1,\ldots,\lambda$}{
			\label{line:third}Sample $\ell^{(i)}$ from $\Bin(n,c_2r/n)$, 
			create $y^{(i)} \assign \mut_{\ell^{(i)}}(x)$, and
			evaluate $f(y^{(i)})$\;
		 }
		$x^* \assign \arg\max\{f(y^{(1)}), \ldots, f(y^{(\lambda)})\}$ (ties broken u.a.r.)\;
		\lIf{$f(x^*)\ge f(x)$}{$x \assign x^*$}
		Perform one of the following two actions with prob. $1/2:$
        \begin{itemize}
           \item replace $r$ with the mutation rate that $x^*$ has been created with;
           \item replace $r$ with either $c_1r$ or $c_2r$ equiprobably.
        \end{itemize}  
        $r \assign \min\{\max\{2, r\}, n/4\}$\;
	}
\caption{The \threerateC with adaptive mutation rates and three subpopulations}
\label{alg:threerate}
\end{algorithm2e}

As we have seen in the previous sections, the \rate performs worse than the \AB for large population sizes. To ensure that this is not caused by a poor tuning of the \rate, we consider the \threerateC on $\lambda = 1{,}600$ and check 100 different configurations of $(c_1, c_2)$ hyper-parameters for this version. The tuning procedure and its results are illustrated in Figure~\ref{fig:multipliers}. First, some random values for $c_1$ and $c_2$ were taken. Then we investigated the areas around values which gave better performance. These results indicated that the best configuration are close to $(c_1=0.7, c_2=1.4)$. It is also worth mentioning that the performance of the \threerateC obtained with the different hyper-parameters differs by up to around 16\%. The relative standard deviation of the tested algorithms is about 1\%. 

Note that we also plot in Figure~\ref{fig:multipliers} the line for which $c_1 = 1/c_2$, because such a functional dependence was used in~\cite{DoerrGWY19}. According to the results of our procedure, the corresponding line does not fit the whole area of good configurations, albeit it does seem to cross some of the best configurations. 

\begin{figure}
\centering
\includegraphics[width=0.5\linewidth]{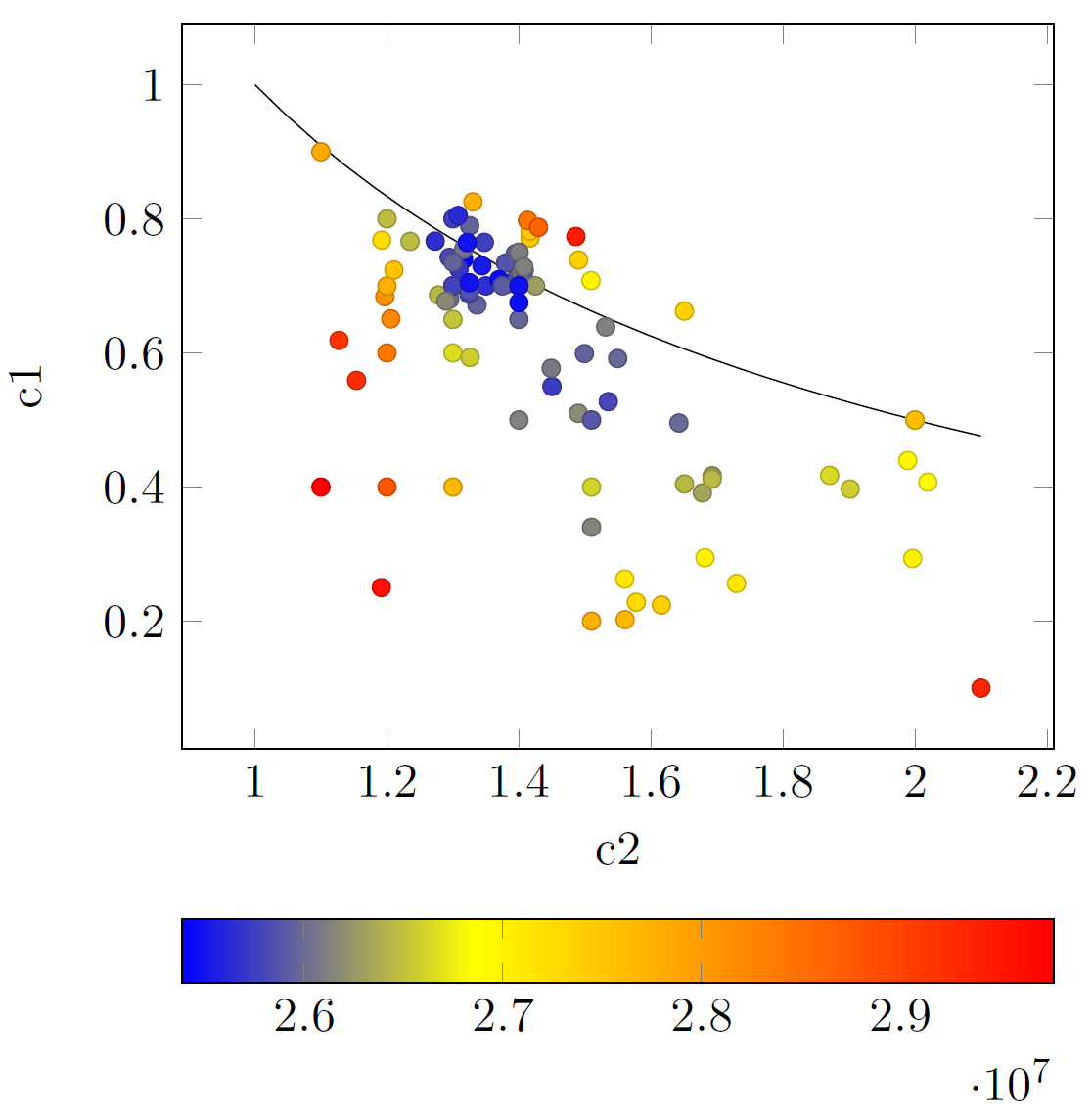}
\caption{Average number of fitness evaluations (shown as color scale) for different hyper-parameters $c_1, c_2$ in \threerateC for $\lambda = 1{,}600$, $n = 100{,}000$.}
\label{fig:multipliers}
\end{figure}

In Figure~\ref{fig:3rate} we plot the performance of the \rateC and the \threerateC using the original hyper-parameters $(c_1 = 2, c_2 = 0.5)$ and using $(c_1 = 1.4, c_2 = 0.7)$. We compare these results to that of the \lea and the \AB for the population size $\lambda = 1{,}600$. 
For the \lea and the \AB the relative standard deviation is around 1\%, while for the \rateC and the \threerateC it is around 2.5\%. 

Figure~\ref{fig:3rate} shows that with the original hyper-parameters the \threerate performs worse than the \rate, while with $c_1 = 1.4, c_2 = 0.7$ they both perform better and very similar to each other. However, the algorithms are still outperformed by the \AB. Therefore, the careful tuning of the hyper-parameters does not influence the performance of the \rateC and the \threerateC drastically compared to the standard choice of $(c_1 = 0.5, c_2 = 2)$, which confirms the observations made in~\cite{DoerrGWY19}, where three values of $c2 = 2.0, 1.5, 1.2$ with a corresponding $c_1 = 1/c_2$ were tested.

\begin{figure}
\centering
\includegraphics[width=0.5\linewidth]{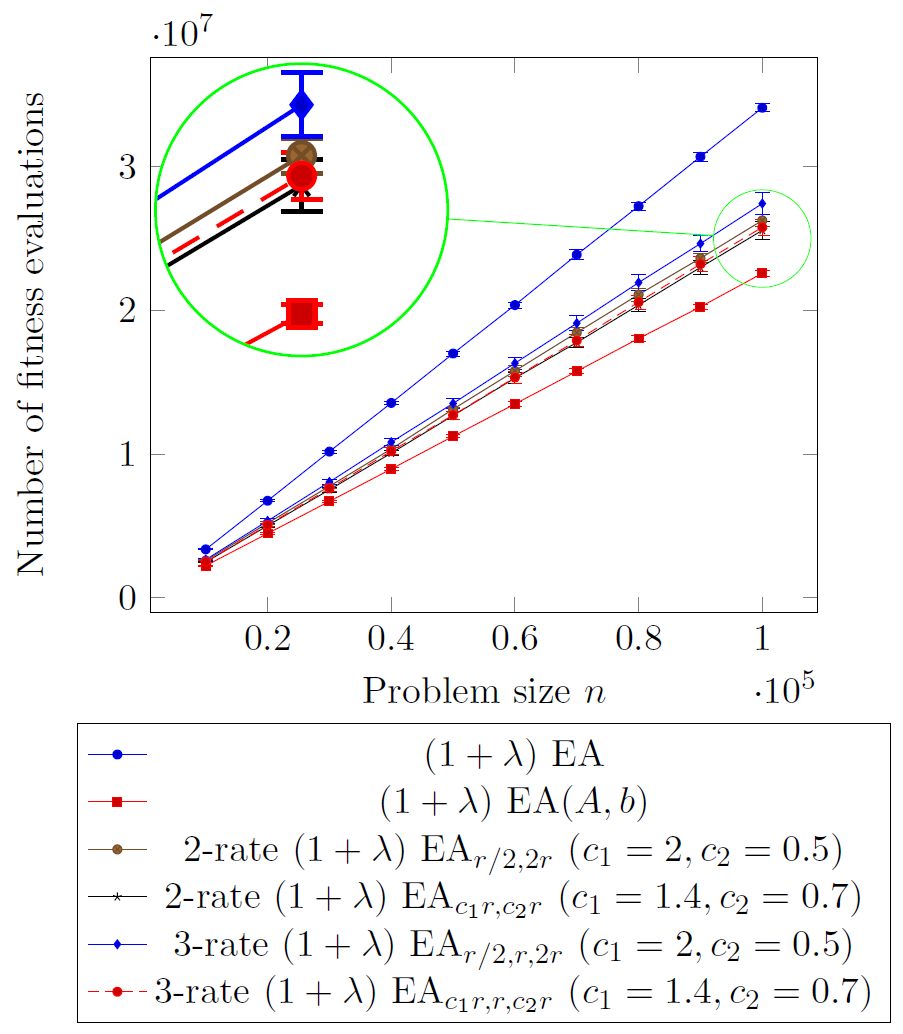}
\caption{Average number of fitness evaluations and its standard deviation for the \rate and the \threerateC with different multipliers compared to the \lea and the \AB for $\lambda = 1{,}600$.}
\label{fig:3rate}
\end{figure}



\section{Conclusions}
\label{sec:conclusions}

We have seen in this work that the ranking of different self-adjusting \lea variants strongly depends on the population size. Both small population sizes and large ones are of interest, as the former provide the best performance in terms of number of fitness evaluations, while the latter are more efficient in terms of generation number, which is useful when calculating fitness in parallel. 

Based on our results, the following conclusions for the \OneMax benchmark problem may be drawn: for small offspring population sizes $5 \leq \lambda \leq \lambda_1$, $\lambda_1$ satisfying $50 \le \lambda_1 < 100$ the \ratelb is the most efficient among the tested algorithms, while for medium population sizes $\lambda_1 \leq \lambda \leq \lambda_2$ with $400 \le \lambda_2 < 800$ the \ABlb performs best, and for large offspring population sizes $\lambda_2 \leq \lambda \leq 3,200$ the \AB with $p_{\min}=1/n$ as the lower bound for the mutation probability should be used. Furthermore, if the evaluation budget is limited and the population size is small, one should pick the algorithms very carefully, as the ranking strongly depends on the budget in this case.

We also confirmed that although a careful selection of the hyper-parameters can improve the performance of the \rate and the \threerate, the default parameters seem to be quite suitably chosen, so that the overall gain of such hyper-parameter tuning seems to be moderate at best. We plan on extending our studies by investigating if similar conclusions can be drawn for the \AB with different values of $A$ and $b$. The results in~\cite{DoerrW18} suggest a rather flat dependence of performance on these hyper-parameters for the \oea, but note that we have considered in our work much larger dimensions and offspring population sizes, so that it is a priori not clear if a similar conclusion holds. We recall from~\cite{Lengler18} that small changes in the parameter setting of an EA can result in drastic performance differences. 

As mentioned in the main part, the consistent ranking of the algorithms makes us feel confident about an extension of our work to a rigorous theoretical running time analysis. Such results would offer more detailed insights into the working principles underlying the diverse rankings, and may help to identify the cut-off points where performances of two algorithms intersect. 

In another important line of research we plan on investigating whether our conclusions for the \onemax problem can be extended to other, more challenging benchmark problems. A promising direction of such extensions is offered by the W-model~\cite{Wmodel}, which allows one to calibrate different features of the optimization problem, such as its ruggedness, the fraction of effective and ``dummy'' variables, the epistasis, neutrality, separability, etc. 

\subsubsection*{Acknowledgments}
Arina Buzdalova was supported by the Government of Russian Federation (Grant 08-08). 
Carola Doerr was supported by a public grant as part of the Investissement d'avenir project, reference ANR-11-LABX-0056-LMH, and by the Paris Ile-de-France Region. We also acknowledge support from COST Action CA15140 'Improving Applicability of Nature-Inspired Optimisation by Joining Theory and Practice (ImAppNIO)' supported by the European Cooperation in Science and Technology.

}

\newcommand{\etalchar}[1]{$^{#1}$}

\newpage
\onecolumn

\appendix

\section{Average Optimization Times and Number of Generations}
\label{app-plots}

\begin{figure}[!h]
\centering
\includegraphics[width=\linewidth]{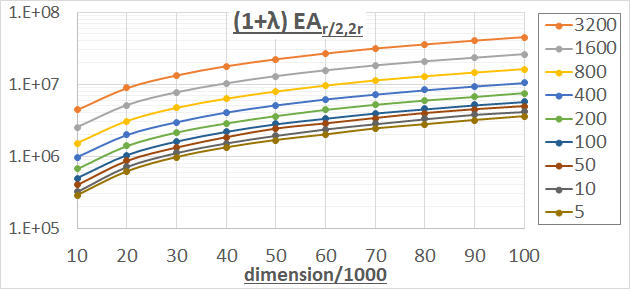}
\caption{Average optimization times for 100 independent runs of the \rate for different values and $5 \le \lambda \le 3,200$ and different problem dimensions $10k \le n \le 100k$. }
\label{fig:RT-2rate-byLambda}
\end{figure}

\begin{figure}[!h]
\centering
\includegraphics[width=\linewidth]{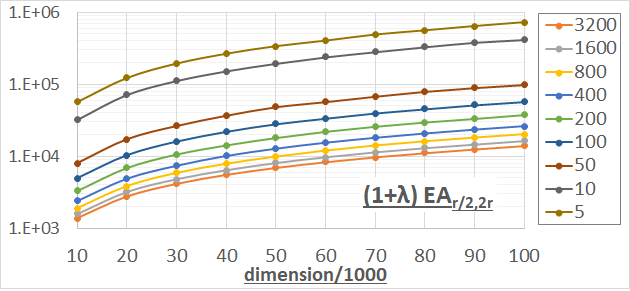}
\caption{Average number of generations for 100 independent runs of the \rate for different values $5 \le \lambda \le 3,200$ and different problem dimensions $10k \le n \le 100k$. }
\label{fig:RT-2rate-byLambda-gen}
\end{figure}

\begin{figure}[!h]
\centering
\includegraphics[width=\linewidth]{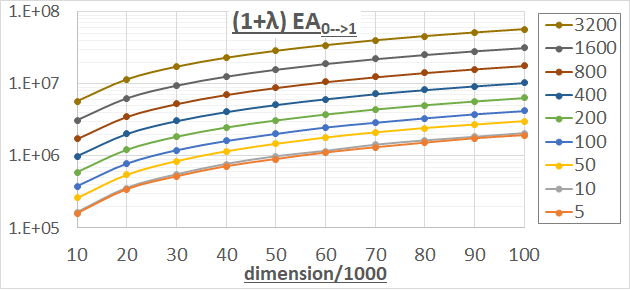}
\caption{Average optimization times for 100 independent runs of the \oplshift for different values $5 \le \lambda \le 3,200$ and different problem dimensions $10k \le n \le 100k$, normalized by $n \ln n$. }
\label{fig:RT-opl-byLambda}
\end{figure}

\begin{figure}[!h]
\centering
\includegraphics[width=\linewidth]{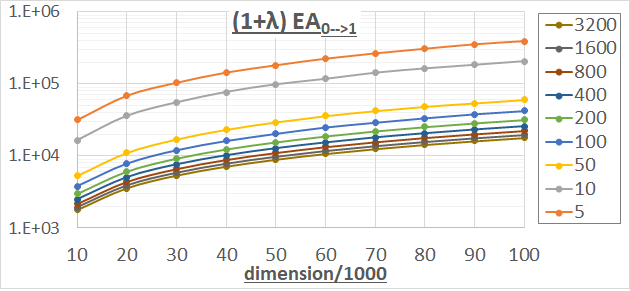}
\caption{Average number of generations for 100 independent runs of the \oplshift for different values $5 \le \lambda \le 3,200$ and different problem dimensions $10k \le n \le 100k$, normalized by $n \ln n$. }
\label{fig:RT-opl-byLambda-gen}
\end{figure}


\begin{figure}
\centering
\includegraphics[width=\linewidth]{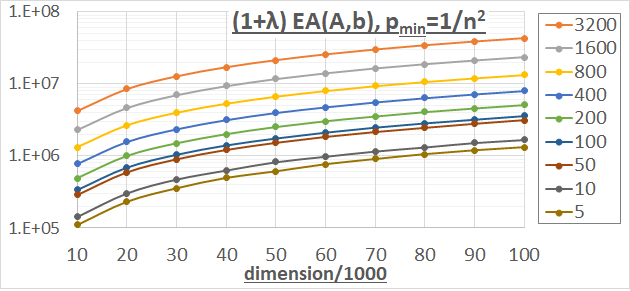}
\caption{Average optimization times for 100 independent runs of the \ABlb for different values $5 \le \lambda \le 3,200$ and different problem dimensions $10k \le n \le 100k$. }
\label{fig:RT-AB-byLambda-n2}
\end{figure}

\begin{figure}
\centering
\includegraphics[width=\linewidth]{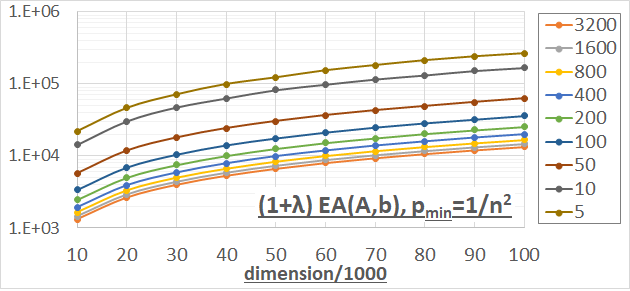}
\caption{Average number of generations for 100 independent runs of the \ABlb for different values $5 \le \lambda \le 3,200$ and different problem dimensions $10k \le n \le 100k$. }
\label{fig:RT-AB-byLambda-gen-n2}
\end{figure}

\begin{figure}[!h]
\centering
\includegraphics[width=\linewidth]{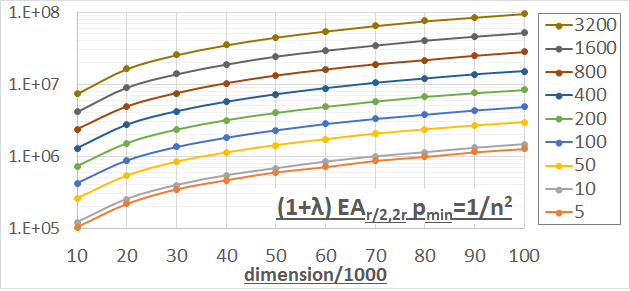}
\caption{Average optimization times for 100 independent runs of the \ratelb for different values $5 \le \lambda \le 3,200$ and different problem dimensions $10k \le n \le 100k$. }
\label{fig:RT-2rate-byLambda-n2}
\end{figure}

\begin{figure}[!h]
\centering
\includegraphics[width=\linewidth]{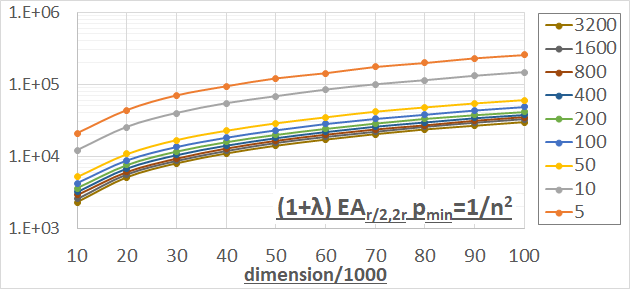}
\caption{Average number of generations for 100 independent runs of the \ratelb for different values $5 \le \lambda \le 3,200$ and different problem dimensions $10k \le n \le 100k$. }
\label{fig:RT-2rate-byLambda-gen-n2}
\end{figure}


\newpage 

{\footnotesize{
\begin{sidewaystable}
\begin{tabular}{rr|rr|rr|rr|rr|rr}
$n$ & $\lambda$ & \multicolumn{2}{|c}{\oplshift}  & \multicolumn{2}{|c}{2-rate $(1 / n)$} & \multicolumn{2}{|c}{Ab $(1 / n)$}   &\multicolumn{2}{|c}{2-rate $(1 / n^2)$ } & \multicolumn{2}{|c}{Ab $(1/n^2)$}  \\
 &  & avg. & r.dev. & avg. & r.dev. & avg. & r.dev. & avg. & r.dev. & avg. & r.dev. \\
\hline
 10,000 	 & 	 1 	 & 	 147,008 	 & 	14.8\%	 & 	 177,568 	 & 	14.8\%	 & 	 148,182 	 & 	14.1\%	 & 	 90,459 	 & 	13.0\%	 & 	 91,563 	 & 	15.6\%	 \\ 
	 & 	 5 	 & 	 31,738 	 & 	17.5\%	 & 	 56,940 	 & 	17.3\%	 & 	 34,514 	 & 	13.0\%	 & 	 20,657 	 & 	13.0\%	 & 	 21,999 	 & 	13.6\%	 \\ 
	 & 	 10 	 & 	 16,373 	 & 	13.1\%	 & 	 32,054 	 & 	15.3\%	 & 	 20,662 	 & 	9.2\%	 & 	 12,036 	 & 	11.1\%	 & 	 14,252 	 & 	7.9\%	 \\ 
	 & 	 50 	 & 	 5,254 	 & 	7.5\%	 & 	 8,029 	 & 	12.9\%	 & 	 6,855 	 & 	6.7\%	 & 	 5,198 	 & 	8.7\%	 & 	 5,740 	 & 	3.7\%	 \\ 
	 & 	 100 	 & 	 3,790 	 & 	4.6\%	 & 	 4,922 	 & 	9.4\%	 & 	 3,824 	 & 	5.8\%	 & 	 4,211 	 & 	9.7\%	 & 	 3,360 	 & 	4.0\%	 \\ 
	 & 	 200 	 & 	 2,955 	 & 	3.6\%	 & 	 3,323 	 & 	8.2\%	 & 	 2,647 	 & 	3.7\%	 & 	 3,603 	 & 	10.5\%	 & 	 2,447 	 & 	2.8\%	 \\ 
	 & 	 400 	 & 	 2,486 	 & 	2.1\%	 & 	 2,420 	 & 	5.7\%	 & 	 2,001 	 & 	2.4\%	 & 	 3,227 	 & 	11.8\%	 & 	 1,943 	 & 	1.8\%	 \\ 
	 & 	 800 	 & 	 2,164 	 & 	1.1\%	 & 	 1,903 	 & 	3.8\%	 & 	 1,633 	 & 	1.8\%	 & 	 2,932 	 & 	13.8\%	 & 	 1,639 	 & 	1.3\%	 \\ 
	 & 	 1,600 	 & 	 1,945 	 & 	0.8\%	 & 	 1,582 	 & 	2.1\%	 & 	 1,393 	 & 	1.2\%	 & 	 2,576 	 & 	11.1\%	 & 	 1,450 	 & 	0.8\%	 \\ 
	 & 	 3,200 	 & 	 1,776 	 & 	0.6\%	 & 	 1,379 	 & 	1.8\%	 & 	 1,227 	 & 	0.8\%	 & 	 2,305 	 & 	12.3\%	 & 	 1,323 	 & 	0.8\%	 \\ 
 \hline 																							
 20,000 	 & 	 1 	 & 	 311,947 	 & 	12.6\%	 & 	 399,856 	 & 	15.1\%	 & 	 318,741 	 & 	14.6\%	 & 	 195,946 	 & 	11.7\%	 & 	 192,649 	 & 	12.8\%	 \\ 
	 & 	 5 	 & 	 68,654 	 & 	14.3\%	 & 	 122,178 	 & 	15.7\%	 & 	 73,448 	 & 	13.8\%	 & 	 43,440 	 & 	9.4\%	 & 	 46,216 	 & 	10.1\%	 \\ 
	 & 	 10 	 & 	 35,726 	 & 	11.1\%	 & 	 71,314 	 & 	11.0\%	 & 	 44,458 	 & 	10.1\%	 & 	 25,398 	 & 	10.1\%	 & 	 30,310 	 & 	8.4\%	 \\ 
	 & 	 50 	 & 	 10,989 	 & 	8.0\%	 & 	 17,278 	 & 	10.5\%	 & 	 14,178 	 & 	5.8\%	 & 	 10,728 	 & 	6.7\%	 & 	 11,840 	 & 	4.5\%	 \\ 
	 & 	 100 	 & 	 7,803 	 & 	5.4\%	 & 	 10,324 	 & 	9.6\%	 & 	 7,765 	 & 	5.8\%	 & 	 8,737 	 & 	7.3\%	 & 	 6,890 	 & 	3.7\%	 \\ 
	 & 	 200 	 & 	 6,052 	 & 	3.5\%	 & 	 6,923 	 & 	7.9\%	 & 	 5,381 	 & 	4.0\%	 & 	 7,554 	 & 	9.8\%	 & 	 4,978 	 & 	3.1\%	 \\ 
	 & 	 400 	 & 	 5,044 	 & 	2.1\%	 & 	 4,924 	 & 	4.5\%	 & 	 4,054 	 & 	2.8\%	 & 	 6,827 	 & 	11.3\%	 & 	 3,924 	 & 	1.9\%	 \\ 
	 & 	 800 	 & 	 4,375 	 & 	1.5\%	 & 	 3,837 	 & 	3.6\%	 & 	 3,286 	 & 	1.6\%	 & 	 6,093 	 & 	9.1\%	 & 	 3,304 	 & 	1.2\%	 \\ 
	 & 	 1,600 	 & 	 3,921 	 & 	0.9\%	 & 	 3,203 	 & 	2.3\%	 & 	 2,791 	 & 	1.1\%	 & 	 5,625 	 & 	10.8\%	 & 	 2,913 	 & 	0.9\%	 \\ 
	 & 	 3,200 	 & 	 3,574 	 & 	0.5\%	 & 	 2,762 	 & 	1.6\%	 & 	 2,448 	 & 	0.5\%	 & 	 5,112 	 & 	11.8\%	 & 	 2,650 	 & 	0.6\%	 \\ 
 \hline 																							
 30,000 	 & 	 1 	 & 	 487,226 	 & 	10.6\%	 & 	 630,449 	 & 	17.1\%	 & 	 503,393 	 & 	12.7\%	 & 	 304,196 	 & 	11.1\%	 & 	 310,610 	 & 	13.3\%	 \\ 
	 & 	 5 	 & 	 103,709 	 & 	10.3\%	 & 	 192,890 	 & 	14.6\%	 & 	 114,129 	 & 	13.1\%	 & 	 69,405 	 & 	10.8\%	 & 	 71,197 	 & 	9.0\%	 \\ 
	 & 	 10 	 & 	 54,922 	 & 	12.6\%	 & 	 111,448 	 & 	14.0\%	 & 	 67,800 	 & 	8.6\%	 & 	 39,620 	 & 	8.9\%	 & 	 47,255 	 & 	6.9\%	 \\ 
	 & 	 50 	 & 	 16,963 	 & 	7.5\%	 & 	 26,418 	 & 	10.5\%	 & 	 21,820 	 & 	6.9\%	 & 	 16,738 	 & 	7.6\%	 & 	 17,956 	 & 	4.0\%	 \\ 
	 & 	 100 	 & 	 11,987 	 & 	4.9\%	 & 	 16,021 	 & 	8.9\%	 & 	 12,149 	 & 	6.2\%	 & 	 13,445 	 & 	6.7\%	 & 	 10,397 	 & 	4.2\%	 \\ 
	 & 	 200 	 & 	 9,200 	 & 	3.7\%	 & 	 10,656 	 & 	7.9\%	 & 	 8,180 	 & 	4.3\%	 & 	 11,681 	 & 	7.8\%	 & 	 7,477 	 & 	2.7\%	 \\ 
	 & 	 400 	 & 	 7,617 	 & 	2.3\%	 & 	 7,438 	 & 	4.2\%	 & 	 6,152 	 & 	3.0\%	 & 	 10,475 	 & 	8.8\%	 & 	 5,882 	 & 	1.8\%	 \\ 
	 & 	 800 	 & 	 6,588 	 & 	1.2\%	 & 	 5,844 	 & 	3.2\%	 & 	 4,957 	 & 	1.8\%	 & 	 9,345 	 & 	8.5\%	 & 	 4,967 	 & 	1.2\%	 \\ 
	 & 	 1,600 	 & 	 5,885 	 & 	0.7\%	 & 	 4,823 	 & 	2.0\%	 & 	 4,190 	 & 	1.1\%	 & 	 8,643 	 & 	12.8\%	 & 	 4,377 	 & 	0.7\%	 \\ 
	 & 	 3,200 	 & 	 5,371 	 & 	0.5\%	 & 	 4,142 	 & 	1.4\%	 & 	 3,675 	 & 	0.7\%	 & 	 7,978 	 & 	11.7\%	 & 	 3,981 	 & 	0.6\%	 \\ 
 \hline 
		\end{tabular}
\caption{Raw Data for Plots in Section~\ref{app-plots}. Number of generations averaged by 100 runs and its standard deviation for \oplshift, \rate, \AB, \ratelb, \ABlb. Shift mutation operator is used in all algorithms, avg.=average, r.dev.=relative standard deviation.}
\label{app:raw-data}
\end{sidewaystable}
}}

{\footnotesize{
\begin{sidewaystable}
\begin{tabular}{rr|rr|rr|rr|rr|rr}
$n$ & $\lambda$ & \multicolumn{2}{|c}{\oplshift}  & \multicolumn{2}{|c}{2-rate $(1 / n)$} & \multicolumn{2}{|c}{Ab $(1 / n)$}   &\multicolumn{2}{|c}{2-rate $(1 / n^2)$ } & \multicolumn{2}{|c}{Ab $(1/n^2)$}  \\
 &  & avg. & r.dev. & avg. & r.dev. & avg. & r.dev. & avg. & r.dev. & avg. & r.dev. \\
\hline																							
 40,000 	 & 	 1 	 & 	 679,092 	 & 	11.9\%	 & 	 856,856 	 & 	13.3\%	 & 	 674,918 	 & 	11.3\%	 & 	 411,103 	 & 	11.4\%	 & 	 426,463 	 & 	13.0\%	 \\ 
	 & 	 5 	 & 	 143,475 	 & 	11.7\%	 & 	 263,636 	 & 	13.1\%	 & 	 157,485 	 & 	12.1\%	 & 	 93,090 	 & 	9.7\%	 & 	 99,060 	 & 	9.8\%	 \\ 
	 & 	 10 	 & 	 76,673 	 & 	10.2\%	 & 	 152,044 	 & 	13.9\%	 & 	 92,633 	 & 	9.9\%	 & 	 54,117 	 & 	8.6\%	 & 	 63,068 	 & 	7.4\%	 \\ 
	 & 	 50 	 & 	 23,156 	 & 	7.0\%	 & 	 36,693 	 & 	10.0\%	 & 	 29,685 	 & 	7.2\%	 & 	 22,507 	 & 	6.2\%	 & 	 24,223 	 & 	4.6\%	 \\ 
	 & 	 100 	 & 	 16,120 	 & 	5.6\%	 & 	 21,882 	 & 	9.9\%	 & 	 16,271 	 & 	6.2\%	 & 	 18,133 	 & 	7.1\%	 & 	 14,026 	 & 	4.4\%	 \\ 
	 & 	 200 	 & 	 12,361 	 & 	3.3\%	 & 	 14,266 	 & 	7.4\%	 & 	 11,013 	 & 	4.4\%	 & 	 15,807 	 & 	7.0\%	 & 	 10,017 	 & 	2.4\%	 \\ 
	 & 	 400 	 & 	 10,194 	 & 	2.0\%	 & 	 10,159 	 & 	5.4\%	 & 	 8,223 	 & 	2.9\%	 & 	 14,167 	 & 	8.3\%	 & 	 7,889 	 & 	1.4\%	 \\ 
	 & 	 800 	 & 	 8,828 	 & 	1.3\%	 & 	 7,824 	 & 	3.0\%	 & 	 6,613 	 & 	1.5\%	 & 	 12,866 	 & 	7.0\%	 & 	 6,623 	 & 	0.8\%	 \\ 
	 & 	 1,600 	 & 	 7,864 	 & 	0.6\%	 & 	 6,427 	 & 	1.9\%	 & 	 5,598 	 & 	1.0\%	 & 	 11,816 	 & 	9.3\%	 & 	 5,842 	 & 	0.7\%	 \\ 
	 & 	 3,200 	 & 	 7,163 	 & 	0.4\%	 & 	 5,541 	 & 	1.3\%	 & 	 4,899 	 & 	0.5\%	 & 	 10,924 	 & 	9.3\%	 & 	 5,314 	 & 	0.5\%	 \\ 
 \hline 																							
 50,000 	 & 	 1 	 & 	 870,823 	 & 	12.9\%	 & 	 1,080,753 	 & 	12.4\%	 & 	 865,991 	 & 	12.5\%	 & 	 528,788 	 & 	14.9\%	 & 	 538,208 	 & 	12.8\%	 \\ 
	 & 	 5 	 & 	 180,681 	 & 	11.8\%	 & 	 334,544 	 & 	14.2\%	 & 	 197,643 	 & 	11.3\%	 & 	 119,915 	 & 	10.5\%	 & 	 122,380 	 & 	9.2\%	 \\ 
	 & 	 10 	 & 	 98,015 	 & 	11.2\%	 & 	 193,599 	 & 	12.5\%	 & 	 117,237 	 & 	9.1\%	 & 	 67,660 	 & 	7.4\%	 & 	 82,358 	 & 	8.8\%	 \\ 
	 & 	 50 	 & 	 29,364 	 & 	7.7\%	 & 	 48,270 	 & 	11.3\%	 & 	 37,081 	 & 	6.1\%	 & 	 28,502 	 & 	5.3\%	 & 	 30,381 	 & 	4.1\%	 \\ 
	 & 	 100 	 & 	 20,257 	 & 	4.7\%	 & 	 27,971 	 & 	9.8\%	 & 	 20,400 	 & 	5.6\%	 & 	 23,060 	 & 	7.1\%	 & 	 17,507 	 & 	3.1\%	 \\ 
	 & 	 200 	 & 	 15,515 	 & 	2.8\%	 & 	 18,045 	 & 	6.8\%	 & 	 13,735 	 & 	3.9\%	 & 	 19,971 	 & 	8.2\%	 & 	 12,590 	 & 	2.4\%	 \\ 
	 & 	 400 	 & 	 12,774 	 & 	2.0\%	 & 	 12,798 	 & 	4.9\%	 & 	 10,306 	 & 	2.4\%	 & 	 17,991 	 & 	9.0\%	 & 	 9,932 	 & 	1.8\%	 \\ 
	 & 	 800 	 & 	 11,033 	 & 	1.2\%	 & 	 9,832 	 & 	3.6\%	 & 	 8,294 	 & 	1.8\%	 & 	 16,457 	 & 	10.3\%	 & 	 8,320 	 & 	1.1\%	 \\ 
	 & 	 1,600 	 & 	 9,835 	 & 	0.7\%	 & 	 8,093 	 & 	2.0\%	 & 	 7,021 	 & 	1.0\%	 & 	 15,261 	 & 	8.9\%	 & 	 7,315 	 & 	0.7\%	 \\ 
	 & 	 3,200 	 & 	 8,956 	 & 	0.5\%	 & 	 6,921 	 & 	1.2\%	 & 	 6,134 	 & 	0.6\%	 & 	 14,005 	 & 	9.9\%	 & 	 6,648 	 & 	0.5\%	 \\ 
 \hline 																							
 60,000 	 & 	 1 	 & 	 1,065,939 	 & 	10.9\%	 & 	 1,321,699 	 & 	12.9\%	 & 	 1,056,875 	 & 	11.4\%	 & 	 665,392 	 & 	12.5\%	 & 	 645,149 	 & 	10.9\%	 \\ 
	 & 	 5 	 & 	 222,882 	 & 	11.6\%	 & 	 398,973 	 & 	10.5\%	 & 	 241,497 	 & 	11.9\%	 & 	 141,691 	 & 	9.4\%	 & 	 153,370 	 & 	9.9\%	 \\ 
	 & 	 10 	 & 	 116,559 	 & 	10.5\%	 & 	 237,965 	 & 	13.4\%	 & 	 143,817 	 & 	9.7\%	 & 	 83,567 	 & 	11.0\%	 & 	 98,172 	 & 	8.9\%	 \\ 
	 & 	 50 	 & 	 35,898 	 & 	8.5\%	 & 	 56,882 	 & 	11.1\%	 & 	 45,142 	 & 	6.3\%	 & 	 34,404 	 & 	6.2\%	 & 	 36,701 	 & 	4.0\%	 \\ 
	 & 	 100 	 & 	 24,894 	 & 	4.8\%	 & 	 33,466 	 & 	7.5\%	 & 	 24,552 	 & 	4.0\%	 & 	 28,244 	 & 	6.2\%	 & 	 21,103 	 & 	3.5\%	 \\ 
	 & 	 200 	 & 	 18,747 	 & 	3.4\%	 & 	 22,022 	 & 	7.2\%	 & 	 16,647 	 & 	3.9\%	 & 	 24,241 	 & 	6.0\%	 & 	 15,183 	 & 	2.6\%	 \\ 
	 & 	 400 	 & 	 15,359 	 & 	2.0\%	 & 	 15,493 	 & 	4.2\%	 & 	 12,456 	 & 	2.6\%	 & 	 21,875 	 & 	8.8\%	 & 	 11,890 	 & 	1.4\%	 \\ 
	 & 	 800 	 & 	 13,297 	 & 	1.3\%	 & 	 11,908 	 & 	3.3\%	 & 	 9,999 	 & 	1.7\%	 & 	 19,981 	 & 	7.8\%	 & 	 9,977 	 & 	0.8\%	 \\ 
	 & 	 1,600 	 & 	 11,828 	 & 	0.8\%	 & 	 9,731 	 & 	2.6\%	 & 	 8,426 	 & 	1.1\%	 & 	 18,467 	 & 	10.9\%	 & 	 8,788 	 & 	0.8\%	 \\ 
	 & 	 3,200 	 & 	 10,758 	 & 	0.4\%	 & 	 8,315 	 & 	1.3\%	 & 	 7,363 	 & 	0.7\%	 & 	 17,024 	 & 	8.5\%	 & 	 7,987 	 & 	0.5\%	 \\ 
 \hline
		\end{tabular}
\caption{Raw Data for Plots in Section~\ref{app-plots}. Number of generations averaged by 100 runs and its standard deviation for \oplshift, \rate, \AB, \ratelb, \ABlb. Shift mutation operator is used in all algorithms, avg.=average, r.dev.=relative standard deviation.}
\label{app:raw-data2}
\end{sidewaystable}
}}

{\footnotesize{
\begin{sidewaystable}
\begin{tabular}{rr|rr|rr|rr|rr|rr}
$n$ & $\lambda$ & \multicolumn{2}{|c}{\oplshift}  & \multicolumn{2}{|c}{2-rate $(1 / n)$} & \multicolumn{2}{|c}{Ab $(1 / n)$}   &\multicolumn{2}{|c}{2-rate $(1 / n^2)$ } & \multicolumn{2}{|c}{Ab $(1/n^2)$}  \\
 &  & avg. & r.dev. & avg. & r.dev. & avg. & r.dev. & avg. & r.dev. & avg. & r.dev. \\
\hline 																							
 70,000 	 & 	 1 	 & 	 1,256,090 	 & 	11.6\%	 & 	 1,575,525 	 & 	12.1\%	 & 	 1,257,547 	 & 	12.8\%	 & 	 763,835 	 & 	12.4\%	 & 	 765,788 	 & 	11.7\%	 \\ 
	 & 	 5 	 & 	 263,995 	 & 	11.8\%	 & 	 484,544 	 & 	11.8\%	 & 	 287,358 	 & 	11.2\%	 & 	 173,293 	 & 	11.6\%	 & 	 181,169 	 & 	10.7\%	 \\ 
	 & 	 10 	 & 	 141,720 	 & 	11.9\%	 & 	 280,866 	 & 	10.5\%	 & 	 168,687 	 & 	9.8\%	 & 	 98,930 	 & 	8.4\%	 & 	 115,841 	 & 	7.6\%	 \\ 
	 & 	 50 	 & 	 42,174 	 & 	6.9\%	 & 	 67,508 	 & 	9.7\%	 & 	 52,244 	 & 	5.5\%	 & 	 41,157 	 & 	6.2\%	 & 	 42,866 	 & 	3.8\%	 \\ 
	 & 	 100 	 & 	 29,002 	 & 	4.6\%	 & 	 39,666 	 & 	8.4\%	 & 	 28,912 	 & 	5.0\%	 & 	 33,200 	 & 	6.3\%	 & 	 24,763 	 & 	3.6\%	 \\ 
	 & 	 200 	 & 	 22,056 	 & 	3.5\%	 & 	 25,905 	 & 	8.6\%	 & 	 19,621 	 & 	4.0\%	 & 	 28,777 	 & 	7.6\%	 & 	 17,649 	 & 	2.1\%	 \\ 
	 & 	 400 	 & 	 18,078 	 & 	2.4\%	 & 	 18,103 	 & 	4.4\%	 & 	 14,510 	 & 	2.7\%	 & 	 25,956 	 & 	8.9\%	 & 	 13,903 	 & 	1.6\%	 \\ 
	 & 	 800 	 & 	 15,550 	 & 	1.3\%	 & 	 14,002 	 & 	3.3\%	 & 	 11,674 	 & 	1.8\%	 & 	 23,607 	 & 	7.1\%	 & 	 11,671 	 & 	0.9\%	 \\ 
	 & 	 1,600 	 & 	 13,817 	 & 	0.6\%	 & 	 11,395 	 & 	1.9\%	 & 	 9,847 	 & 	1.1\%	 & 	 21,763 	 & 	9.0\%	 & 	 10,270 	 & 	0.7\%	 \\ 
	 & 	 3,200 	 & 	 12,579 	 & 	0.4\%	 & 	 9,736 	 & 	1.6\%	 & 	 8,596 	 & 	0.6\%	 & 	 20,283 	 & 	10.0\%	 & 	 9,319 	 & 	0.5\%	 \\ 
\hline																							
 80,000 	 & 	 1 	 & 	 1,464,944 	 & 	11.9\%	 & 	 1,784,910 	 & 	11.9\%	 & 	 1,464,315 	 & 	11.7\%	 & 	 888,194 	 & 	13.1\%	 & 	 899,089 	 & 	12.5\%	 \\ 
	 & 	 5 	 & 	 306,038 	 & 	11.2\%	 & 	 553,383 	 & 	11.1\%	 & 	 325,414 	 & 	9.8\%	 & 	 196,852 	 & 	9.7\%	 & 	 211,511 	 & 	9.9\%	 \\ 
	 & 	 10 	 & 	 162,877 	 & 	10.4\%	 & 	 328,819 	 & 	12.2\%	 & 	 194,256 	 & 	9.4\%	 & 	 112,892 	 & 	7.6\%	 & 	 132,056 	 & 	7.9\%	 \\ 
	 & 	 50 	 & 	 48,433 	 & 	6.9\%	 & 	 79,050 	 & 	11.2\%	 & 	 60,104 	 & 	5.6\%	 & 	 47,120 	 & 	5.7\%	 & 	 48,995 	 & 	3.4\%	 \\ 
	 & 	 100 	 & 	 33,455 	 & 	5.6\%	 & 	 45,309 	 & 	7.1\%	 & 	 33,345 	 & 	4.9\%	 & 	 38,013 	 & 	5.1\%	 & 	 28,374 	 & 	3.3\%	 \\ 
	 & 	 200 	 & 	 25,319 	 & 	4.1\%	 & 	 29,699 	 & 	6.1\%	 & 	 22,312 	 & 	3.9\%	 & 	 33,475 	 & 	8.3\%	 & 	 20,308 	 & 	2.7\%	 \\ 
	 & 	 400 	 & 	 20,620 	 & 	2.2\%	 & 	 20,785 	 & 	6.1\%	 & 	 16,598 	 & 	2.1\%	 & 	 29,709 	 & 	8.2\%	 & 	 15,909 	 & 	1.6\%	 \\ 
	 & 	 800 	 & 	 17,790 	 & 	1.5\%	 & 	 16,014 	 & 	3.7\%	 & 	 13,375 	 & 	1.6\%	 & 	 26,935 	 & 	7.1\%	 & 	 13,330 	 & 	1.0\%	 \\ 
	 & 	 1,600 	 & 	 15,822 	 & 	0.7\%	 & 	 13,010 	 & 	2.1\%	 & 	 11,277 	 & 	1.2\%	 & 	 25,466 	 & 	9.1\%	 & 	 11,724 	 & 	0.6\%	 \\ 
	 & 	 3,200 	 & 	 14,377 	 & 	0.5\%	 & 	 11,096 	 & 	1.1\%	 & 	 9,832 	 & 	0.6\%	 & 	 23,596 	 & 	9.7\%	 & 	 10,652 	 & 	0.5\%	 \\ 
 \hline 																							
 90,000 	 & 	 1 	 & 	 1,644,769 	 & 	12.8\%	 & 	 2,014,968 	 & 	10.4\%	 & 	 1,663,195 	 & 	10.3\%	 & 	 1,018,903 	 & 	12.7\%	 & 	 1,008,852 	 & 	10.4\%	 \\ 
	 & 	 5 	 & 	 351,978 	 & 	11.8\%	 & 	 634,502 	 & 	12.7\%	 & 	 380,506 	 & 	11.4\%	 & 	 228,885 	 & 	9.8\%	 & 	 239,240 	 & 	11.3\%	 \\ 
	 & 	 10 	 & 	 182,743 	 & 	9.6\%	 & 	 379,167 	 & 	11.8\%	 & 	 223,083 	 & 	8.5\%	 & 	 130,161 	 & 	10.8\%	 & 	 151,945 	 & 	6.8\%	 \\ 
	 & 	 50 	 & 	 54,042 	 & 	6.7\%	 & 	 89,176 	 & 	11.0\%	 & 	 67,712 	 & 	5.4\%	 & 	 53,486 	 & 	6.7\%	 & 	 55,810 	 & 	4.1\%	 \\ 
	 & 	 100 	 & 	 37,998 	 & 	5.9\%	 & 	 51,540 	 & 	7.9\%	 & 	 37,983 	 & 	6.6\%	 & 	 43,632 	 & 	5.9\%	 & 	 31,916 	 & 	3.5\%	 \\ 
	 & 	 200 	 & 	 28,529 	 & 	3.7\%	 & 	 33,267 	 & 	6.2\%	 & 	 25,451 	 & 	3.8\%	 & 	 37,658 	 & 	7.6\%	 & 	 22,852 	 & 	2.7\%	 \\ 
	 & 	 400 	 & 	 23,249 	 & 	2.0\%	 & 	 23,480 	 & 	4.4\%	 & 	 18,836 	 & 	2.9\%	 & 	 33,974 	 & 	8.0\%	 & 	 17,956 	 & 	1.6\%	 \\ 
	 & 	 800 	 & 	 20,002 	 & 	1.2\%	 & 	 17,987 	 & 	3.2\%	 & 	 15,023 	 & 	1.6\%	 & 	 31,103 	 & 	9.3\%	 & 	 15,022 	 & 	1.0\%	 \\ 
	 & 	 1,600 	 & 	 17,795 	 & 	0.8\%	 & 	 14,592 	 & 	1.7\%	 & 	 12,653 	 & 	0.9\%	 & 	 28,968 	 & 	9.7\%	 & 	 13,207 	 & 	0.6\%	 \\ 
	 & 	 3,200 	 & 	 16,153 	 & 	0.4\%	 & 	 12,506 	 & 	1.1\%	 & 	 11,064 	 & 	0.6\%	 & 	 26,533 	 & 	9.5\%	 & 	 12,006 	 & 	0.5\%	 \\ 
		\end{tabular}
\caption{Raw Data for Plots in Section~\ref{app-plots}. Number of generations averaged by 100 runs and its standard deviation for \oplshift, \rate, \AB, \ratelb, \ABlb. Shift mutation operator is used in all algorithms, avg.=average, r.dev.=relative standard deviation.}
\label{app:raw-data3}
\end{sidewaystable}
}}

{\footnotesize{
\begin{sidewaystable}
\begin{tabular}{rr|rr|rr|rr|rr|rr}
$n$ & $\lambda$ & \multicolumn{2}{|c}{\oplshift}  & \multicolumn{2}{|c}{2-rate $(1 / n)$} & \multicolumn{2}{|c}{Ab $(1 / n)$}   &\multicolumn{2}{|c}{2-rate $(1 / n^2)$ } & \multicolumn{2}{|c}{Ab $(1/n^2)$}  \\
 &  & avg. & r.dev. & avg. & r.dev. & avg. & r.dev. & avg. & r.dev. & avg. & r.dev. \\	
 \hline 																							
 100,000 	 & 	 1 	 & 	 1,873,666 	 & 	13.4\%	 & 	 2,222,691 	 & 	10.9\%	 & 	 1,860,718 	 & 	13.4\%	 & 	 1,143,933 	 & 	11.7\%	 & 	 1,122,686 	 & 	10.6\%	 \\ 
	 & 	 5 	 & 	 389,110 	 & 	10.2\%	 & 	 716,372 	 & 	13.6\%	 & 	 420,165 	 & 	10.4\%	 & 	 253,747 	 & 	11.7\%	 & 	 264,171 	 & 	8.5\%	 \\ 
	 & 	 10 	 & 	 205,086 	 & 	12.0\%	 & 	 414,043 	 & 	10.4\%	 & 	 248,184 	 & 	9.7\%	 & 	 146,132 	 & 	8.2\%	 & 	 167,244 	 & 	7.0\%	 \\ 
	 & 	 50 	 & 	 60,200 	 & 	6.2\%	 & 	 98,223 	 & 	8.7\%	 & 	 76,772 	 & 	5.6\%	 & 	 59,357 	 & 	5.7\%	 & 	 62,693 	 & 	3.8\%	 \\ 
	 & 	 100 	 & 	 42,151 	 & 	5.0\%	 & 	 57,325 	 & 	6.9\%	 & 	 41,520 	 & 	4.1\%	 & 	 48,659 	 & 	6.6\%	 & 	 35,814 	 & 	4.1\%	 \\ 
	 & 	 200 	 & 	 31,762 	 & 	3.4\%	 & 	 37,588 	 & 	6.8\%	 & 	 28,215 	 & 	3.7\%	 & 	 41,634 	 & 	4.6\%	 & 	 25,431 	 & 	2.2\%	 \\ 
	 & 	 400 	 & 	 25,846 	 & 	2.1\%	 & 	 26,227 	 & 	5.0\%	 & 	 20,900 	 & 	2.6\%	 & 	 37,583 	 & 	8.3\%	 & 	 19,984 	 & 	1.6\%	 \\ 
	 & 	 800 	 & 	 22,229 	 & 	1.2\%	 & 	 20,055 	 & 	3.2\%	 & 	 16,681 	 & 	1.4\%	 & 	 34,965 	 & 	9.3\%	 & 	 16,691 	 & 	0.9\%	 \\ 
	 & 	 1,600 	 & 	 19,744 	 & 	0.7\%	 & 	 16,325 	 & 	2.0\%	 & 	 14,112 	 & 	1.0\%	 & 	 32,494 	 & 	11.6\%	 & 	 14,691 	 & 	0.6\%	 \\ 
	 & 	 3,200 	 & 	 17,956 	 & 	0.4\%	 & 	 13,966 	 & 	1.1\%	 & 	 12,296 	 & 	0.6\%	 & 	 29,796 	 & 	11.1\%	 & 	 13,344 	 & 	0.5\%	
		\end{tabular}
\caption{Raw Data for Plots in Section~\ref{app-plots}. Number of generations averaged by 100 runs and its standard deviation for \oplshift, \rate, \AB, \ratelb, \ABlb. Shift mutation operator is used in all algorithms, avg.=average, r.dev.=relative standard deviation.}
\label{app:raw-data4}
\end{sidewaystable}
}}


\begin{sidewaystable}[h]
\caption{Raw Data for Figure~\ref{fig:3rate}. Average number of fitness evaluations for \rate (2-rate) and \threerate (3-rate) with different hyper-parameters (0.5, 2 and 0.7, 1.4) compared to \lea and \AB (Ab) for $\lambda = 1600.$ Standard mutation operator is used in all algorithms, avg.=average, r.dev.=relative standard deviation.}
\label{tab:3rate}
\begin{tabular}{r|rr|rr|rr|rr|rr|rr}
$n$ & \multicolumn{2}{|c}{\lea}  & \multicolumn{2}{|c}{2-rate (0.5) }  & \multicolumn{2}{|c}{2-rate (0.7) }  & \multicolumn{2}{|c}{3-rate (0.5) }  & \multicolumn{2}{|c}{3-rate (0.7)} & \multicolumn{2}{|c}{Ab}\\
& avg. & r.dev. & avg. & r.dev. & avg. & r.dev. & avg. & r.dev. & avg. & r.dev. & avg. & r.dev.\\
\hline
 10,000 	 & 	 3,362,177 	 & 	1.0\%	 & 	 2,647,488 	 & 	2.4\%	 & 	 2,473,264 	 & 	2.2\%	 & 	 2,645,616 	 & 	3.1\%	 & 	 2,498,816 	 & 	2.5\%	 & 	 2,228,449 	 & 	1.2\%	\\
 20,000 	 & 	 6,744,593 	 & 	1.1\%	 & 	 5,378,496 	 & 	3.2\%	 & 	 5,011,088 	 & 	2.4\%	 & 	 5,341,728 	 & 	3.4\%	 & 	 5,056,128 	 & 	2.5\%	 & 	 4,466,209 	 & 	1.1\%	\\
 30,000 	 & 	 10,167,969 	 & 	1.0\%	 & 	 8,097,168 	 & 	2.5\%	 & 	 7,512,784 	 & 	2.5\%	 & 	 8,055,056 	 & 	2.4\%	 & 	 7,609,648 	 & 	2.5\%	 & 	 6,703,809 	 & 	1.1\%	\\
 40,000 	 & 	 13,556,897 	 & 	0.9\%	 & 	 10,881,344 	 & 	2.4\%	 & 	 10,063,600 	 & 	1.9\%	 & 	 10,825,376 	 & 	2.1\%	 & 	 10,201,920 	 & 	2.6\%	 & 	 8,957,105 	 & 	1.0\%	\\
 50,000 	 & 	 16,995,265 	 & 	1.0\%	 & 	 13,659,776 	 & 	2.8\%	 & 	 12,678,240 	 & 	2.1\%	 & 	 13,523,152 	 & 	2.6\%	 & 	 12,706,512 	 & 	2.2\%	 & 	 11,233,585 	 & 	1.0\%	\\
 60,000 	 & 	 20,369,121 	 & 	1.0\%	 & 	 16,393,136 	 & 	2.4\%	 & 	 15,283,856 	 & 	2.4\%	 & 	 16,322,656 	 & 	2.5\%	 & 	 15,329,392 	 & 	2.6\%	 & 	 13,482,257 	 & 	1.1\%	\\
 70,000 	 & 	 23,874,081 	 & 	1.4\%	 & 	 19,181,872 	 & 	2.5\%	 & 	 17,763,856 	 & 	2.1\%	 & 	 19,106,128 	 & 	2.8\%	 & 	 17,899,920 	 & 	2.2\%	 & 	 15,755,281 	 & 	1.1\%	\\
 80,000 	 & 	 27,254,449 	 & 	1.0\%	 & 	 22,074,384 	 & 	3.0\%	 & 	 20,368,800 	 & 	2.2\%	 & 	 21,933,152 	 & 	2.5\%	 & 	 20,575,680 	 & 	2.4\%	 & 	 18,043,073 	 & 	1.2\%	\\
 90,000 	 & 	 30,697,777 	 & 	1.0\%	 & 	 24,806,320 	 & 	2.6\%	 & 	 22,990,288 	 & 	2.1\%	 & 	 24,648,208 	 & 	2.3\%	 & 	 23,221,920 	 & 	2.3\%	 & 	 20,245,025 	 & 	0.9\%	\\
 100,000 	 & 	 34,105,201 	 & 	0.8\%	 & 	 27,558,720 	 & 	2.4\%	 & 	 25,569,168 	 & 	2.4\%	 & 	 27,436,784 	 & 	2.7\%	 & 	 25,784,448 	 & 	2.1\%	 & 	 22,579,393 	 & 	1.0\%	
\end{tabular}
\end{sidewaystable}


\end{document}